\documentclass[sigconf,9pt]{acmart}
\pdfoutput=1

\setcopyright{none}
\renewcommand\footnotetextcopyrightpermission[1]{}

\usepackage{verbatim}
\usepackage{underscore}
\usepackage{graphics} 
\usepackage{epsfig} 
\usepackage{epstopdf}
\usepackage{subfigure}
\usepackage{balance}
\usepackage{comment}
\usepackage{booktabs}
\usepackage{amsmath}
\usepackage{amsfonts}
\usepackage[english]{babel}
\usepackage[utf8]{inputenc}
\usepackage{algorithm}
\usepackage[noend]{algpseudocode}
\usepackage{bm}
\usepackage{colortbl}
\usepackage{tabularx}
\usepackage{color}
\usepackage{xspace}
\usepackage{graphicx}    
\usepackage{url}         
\usepackage{algorithm,algorithmicx}
\usepackage{placeins}
\usepackage{multirow}
\usepackage[skip=0pt]{caption}
\usepackage{enumitem}
\usepackage{leading}
\usepackage{pgfplots}
\usepackage{balance}
\floatname{algorithm}{Algorithm}

\usepackage{etoolbox}
\newcommand{\zerodisplayskips}{%
  \setlength{\abovedisplayskip}{2pt}%
  \setlength{\belowdisplayskip}{2pt}%
  \setlength{\abovedisplayshortskip}{2pt}%
  \setlength{\belowdisplayshortskip}{2pt}}
\appto{\normalsize}{\zerodisplayskips}
\appto{\small}{\zerodisplayskips}
\appto{\footnotesize}{\zerodisplayskips}

\newcommand{\SystemName}{FreeGaze\xspace}

\copyrightyear{2022} 
\acmYear{2022} 

\begin{document}
\title{\SystemName: Resource-efficient Gaze Estimation via Frequency Domain Contrastive Learning}

\author{Lingyu Du}
\affiliation{%
  \institution{Delft University of Technology}
    \city{Delft}
   \country{The Netherlands}
}
\email{lingyu.du@tudelft.nl}

\author{Guohao Lan}
\affiliation{%
  \institution{Delft University of Technology}
  \city{Delft}
   \country{The Netherlands}
}
\email{g.lan@tudelft.nl}

\begin{abstract}
Gaze estimation is of great importance to many scientific fields and daily applications, ranging from fundamental research in cognitive psychology to attention-aware mobile systems. While recent advancements in deep learning have yielded remarkable successes in building highly accurate gaze estimation systems, the associated high computational cost and the reliance on large-scale labeled gaze data for supervised learning place challenges on the practical use of
existing solutions. To move beyond these limitations, we present \SystemName, a resource-efficient framework for unsupervised gaze representation learning. \SystemName incorporates \textit{the frequency domain gaze estimation} and \textit{the contrastive gaze representation learning} in its design. The former significantly alleviates the computational burden in both system calibration and gaze estimation, and dramatically reduces the system latency; while the latter overcomes the data labeling hurdle of existing supervised learning-based counterparts, and ensures efficient gaze representation learning in the absence of gaze label. Our evaluation on two gaze estimation datasets shows that \SystemName can achieve comparable gaze estimation accuracy with existing supervised learning-based approach, while enabling up to 6.81 and 1.67 times speedup in system calibration and gaze estimation, respectively. 
\end{abstract}

\maketitle
	
\section{Introduction}
\label{sec:introduction}

Human gaze indicates where the subject is looking at, and is widely considered as an important non-verbal communication cue that expresses the subject's desires, needs, cognitive stages, and interpersonal relations~\cite{hansen2009eye}. The estimation of human gaze, thus, is of great importance to a wide range of scientific fields~\cite{eckstein2017beyond}. For instance, in cognitive psychology, 
human gaze provides unique insights into the understanding of mental disorders~\cite{shishido2019application,duan2019dataset}, personality traits~\cite{berkovsky2019detecting}, cognitive workload~\cite{kosch2018your,fridman2018cognitive}, and visual perception~\cite{de2019individual,najemnik2005optimal}. 

Beyond the benefits to fundamental science, there is also an unprecedented demand for eye tracking in mobile systems. In the context of virtual and augmented reality (AR/VR), accurate and fast gaze estimation is crucial for ensuring immersive user experiences. Examples such as eye tracking-based dynamic scene adaptation~\cite{marwecki2019mise} and foveated rendering~\cite{patney2016towards,kim2019foveated}. 
Outside the field of AR/VR, general-purpose RGB cameras, such as those embedded in smartphones~\cite{valliappan2020accelerating}, tablets~\cite{wood2014eyetab}, and webcams~\cite{sugano2016aggregaze}, can also be used to capture facial images of the user and enable appearance-based gaze estimation systems~\cite{cheng2021appearance}. In fact, gaze has been used 
to improve user perceived quality in webpage loading~\cite{kelton2017improving}, to recognize daily human activity~\cite{lan2020gazegraph,srivastava2018combining}, and to ensure friendly human-robot interaction~\cite{gillet2021robot,weldon2021exploring}. 
Despite the numerous benefits, there are two practical challenges that hinder the wide adoption of existing gaze estimation solutions. 

\textbf{Challenges in collecting large-scale labeled gaze data.} Similar to the other computer vision tasks, the advancements in convolutional neural networks (CNN) have contributed significantly to the success in gaze estimation~\cite{kassner2014pupil,cheng2021appearance,zhang2020eth,krafka2016eye,zhang2018training,zhang2017mpiigaze,huynh2021imon}. However, like many CNN-based systems, existing gaze estimation solutions rely on supervised learning, and thus, their performance is largely depending on the availability of a large-scale, well-labeled training dataset. Unfortunately, \textit{collecting gaze data with accurately annotated labels} is a labor-intensive and impractical process that requires controlled and highly sophisticated setups and subject recruitment~\cite{kellnhofer2019gaze360,zhang2017mpiigaze,zhang2020eth,newman2020turkeyes}. For instance, existing gaze datasets are collected using expensive, high-resolution cameras~\cite{kellnhofer2019gaze360,zhang2017mpiigaze,zhang2020eth}, or relying on dedicated crowdsourcing 
platforms~\cite{krafka2016eye,newman2020turkeyes}. What makes the gaze data collection more challenging is that the accuracy of gaze annotation is highly influenced by the user's fixation ability and cooperation during the collection process~\cite{openeds}. For example, due to the subconscious eye movements, the participant may not gaze at the ``ground-truth'' point that she supposed to fixed on, and thus, leads to incorrect gaze labels and dirty data. Therefore, while complex CNN models are worthwhile for ensuring high gaze estimation accuracy, we need an efficient and inexpensive solution to train the system, in the absence of gaze label.

\textbf{Challenges in system latency.} 
The adoption of complex neural networks also introduces high 
system latency in both calibration and gaze estimation stages, which places limitations on the practical use of existing solutions. 

First, to eliminate the impact of subject diversity on the gaze estimation accuracy, all existing solutions need to perform the subject-specific calibration before the system's first use~\cite{hansen2009eye,cheng2021appearance,sugano2015self,lanata2015robust,linden2019learning}. The calibration fine-tunes the pre-trained gaze estimation model by a small set of labeled gaze data collected from the targeted subject \textit{on the fly}. Thus, when complex neural networks are employed~\cite{cheng2021appearance}, the online calibration becomes a time-consuming process and affects the user's experience. 
Indeed, our {measurement study} (Section~\ref{subsec:timeEfficiency}) indicates that a ResNet-18 based solution (which is 32-layers shallower than the ResNet-50 based design that is widely adopted by existing solutions~\cite{zhang2020eth}) takes 10.2 minutes for calibration, despite a high-end GeForce RTX 3080Ti GPU is used. Note that, calibration is also inevitable for high-end model-based eye tracking systems, e.g., Tobii Pro~\cite{tobiiCalibration}, Pupil Lab~\cite{pupilLabCalibration}, and VIVE Pro Eye VR headset~\cite{vevoCalibration}, though their calibration is performed in a slightly different way.

    
Second, mobile systems usually have a {constrained time budget for run-time gaze estimation}. For instance, foveated rendering~\cite{patney2016towards,kim2019foveated} requires a low gaze estimation latency to render a high-quality image precisely in the user's foveal vision on time, 
even when the user is making fast eye movements such as saccades~\cite{albert2017latency}. 
Similarly, eye tracking-based psychological applications~\cite{mele2012gaze} require a high tracking rate to capture fine-grained cognitive contexts. Thus, in practical application scenario, gaze estimation systems often face the requirements of low inference latency.


To address these challenges, we present \SystemName, a resource-efficient framework for unsupervised gaze representation learning. At the core of \SystemName are the \textit{frequency domain gaze estimation} and the \textit{contrastive gaze representation learning}. The former significantly alleviates the computational burden in both calibration and estimation, and dramatically reduces the system latency; while the latter overcomes the data labeling hurdle of existing supervised learning-based counterparts, and ensures efficient gaze representation learning without gaze label. These capabilities are made possible by a suite of novel techniques devised in this work. 

First, existing gaze estimation systems take RGB images as inputs. To reduce the system latency, a straightforward way is to aggressively reduce the input size of the neural networks. However, simply compressing or downsampling the RGB image will destroy the perceptual information it contains and leads to poor estimation performance. To resolve this challenge, we introduce the frequency domain gaze estimation (Section~\ref{subsec:DCT}), which employs the discrete cosine transform (DCT) to convert the original RGB images to the frequency domain, and takes the DCT coefficients as inputs for gaze estimation. Moreover, we exploit the \textit{spectral compaction property}~\cite{rao2014discrete} of the DCT, which concentrates the critical content-defining information of the image in the low-end of the frequency spectrum, while putting the trivial and noise signals in the high-frequency end~\cite{wallace1991jpeg}. This motivates us to aggressively compact the essential perceptual information in the RGB image into a few DCT coefficients in the low-frequency domain. In fact, as demonstrated in Section~\ref{sec:evaluation}, when comparing with conventional RGB-based solutions, the proposed frequency domain gaze estimation achieves up to 6.81 and 1.67 times speedup in calibration and gaze estimation, respectively. 

Second, to overcome the data labeling hurdle of existing supervised gaze estimation systems~\cite{kassner2014pupil,zhang2020eth,krafka2016eye,zhang2018training,zhang2017mpiigaze,huynh2021imon}, we propose a contrastive learning (CL)-based framework (Section~\ref{subsec:CL}) that leverages unlabeled facial images for gaze representation learning. 
Although various CL-based unsupervised learning methods have been proposed for general purpose visual tasks, such as image classification~\cite{chen2020simple,dosovitskiy2014discriminative} and object detection~\cite{xie2021detco}, they are ill-suited for gaze estimation, as they focus on learning general representations that are more related to the appearance and the identity of the subjects. In fact, as demonstrated in Sections~\ref{sec:background} and~\ref{sec:evaluation}, conventional CL approach, i.e., SimCLR~\cite{chen2020simple}, can lead to poor performance in gaze representation learning and high gaze estimation error. To resolve this challenge, we devise a set of techniques for contrastive gaze representation learning. Specifically, we introduce the \textit{subject-specific negative pair sampling strategy} (Section~\ref{subsec:subjectSpecificNegativePairSampling}) to encourage the learning of gaze-related features, and design the \textit{gaze-specific data augmentation} (Section~\ref{subsec:dataAugmentation}) to ensure the gaze consistence during the contrastive learning. As demonstrated 
(Section~\ref{subsec:gazeEstimationPerformance}), the two techniques lead to significant improvements in gaze estimation when comparing with the conventional unsupervised method~\cite{chen2020simple}. 

Our major contributions are summarized as follows:
\vspace{-0.05in}
\begin{itemize}[leftmargin=*, wide, labelwidth=!, labelindent=0pt]
\item We introduce the frequency domain gaze estimation that exploits the spectral compaction property of the discrete cosine transform to aggressively compress the inputs of gaze estimation systems and achieves significant speedup in both calibration and gaze estimation stages. To our best knowledge, this is the first work that leverages the DCT coefficients for gaze estimation.

\item We propose the frequency domain contrastive gaze representation learning framework that overcomes the data labeling hurdle of existing supervised learning-based counterparts. We devise the \textit{subject-specific negative pair sampling strategy} and the \textit{gaze-specific data augmentation} to ensure efficient self-supervised gaze representation learning.

\item We conduct a comprehensive evaluation of \SystemName on two state-of-the-art gaze estimation datasets. 
The results demonstrate the effectiveness of \SystemName in learning gaze representations in the compressed frequency domain. Specifically, \SystemName reduces the angular error of the conventional CL-based method by 5.4$^\circ$ and 4.6$^\circ$ on average over the two datasets, respectively. When comparing with the RGB supervised learning-based method, which takes 37K and 30K labeled images for training, \SystemName achieves comparable results, i.e., the performance gap is only 1.1$^\circ$ and 1.6$^\circ$ on average over the two datasets, while enabling up to 6.81 and 1.67 times speedup in calibration and gaze estimation, respectively. 

\end{itemize}

\vspace{-0.05in}
The rest of the paper is organized as follows. We review related work in Section~\ref{sec:related_work}. We introduce the motivation and challenges in Section~\ref{sec:background}. The design details of \SystemName are given in Section~\ref{sec:contrastiveLearning}. We evaluate our work in Section~\ref{sec:evaluation} followed by the conclusion in Section~\ref{sec:conlusion}. 

\section{Related Work}
\label{sec:related_work}







\subsection{Gaze Estimation} 


Existing gaze estimation systems can be classified into model-based and appearance-based methods. Model-based estimation systems extract features from eye images to fit a geometric model that predicts gaze direction~\cite{hansen2009eye}. However, they require 
dedicated eye tracking camera~\cite{kassner2014pupil,nakazawa2012point,guestrin2006general} to capture near-eye images for the estimation, which hinders their pervasive adoption. By contrast, appearance-based solutions rely on ordinary facial or eye images captured by general-purpose RGB cameras, such as those embedded in smartphones~\cite{valliappan2020accelerating,krafka2016eye}, tablets~\cite{wood2014eyetab,zhang2018training}, and public displays~\cite{sugano2016aggregaze,zhang2015appearance}, which makes them promising for a wide range of everyday settings. The latest appearance-based gaze estimation systems benefit from the representation learning capability of deep neural networks, and have shown promising performance that significantly surpassed their model-based counterparts~\cite{cheng2021appearance}. Indeed, over the last couple of years, a plethora of deep learning-based gaze estimation solutions have been proposed~\cite{valliappan2020accelerating,krafka2016eye,zhang2018training,zhang2017mpiigaze,zhang2020eth,huynh2021imon}. 
For instance, Zhang et al.~\cite{zhang2015appearance} took eye images and head pose information estimated from facial images as inputs to train a multi-modal CNN for gaze estimation. They also showed that adopting a deeper neural networks, i.e., the ResNet-50, can further improve the estimation accuracy~\cite{zhang2017mpiigaze,zhang2020eth}. 


Despite the promise of recent deep learning-based solutions, they rely on supervised learning and require a large labeled gaze dataset for training, e.g., 210K images for the MPIIGaze~\cite{zhang2017mpiigaze} and 750K images for the ETH-XGaze~\cite{zhang2020eth}, which is complex and expensive to be collected and annotated. By contrast, \SystemName achieves gaze estimation in an unsupervised manner. 
Moreover, by taking frequency domain representations as inputs, \SystemName achieves up to 6.81 and 1.67 times speedup in system calibration and inference, respectively, when comparing with the conventional RGB-based gaze estimation solution.

\subsection{Unsupervised Representation Learning}

Our work is also related to existing efforts in unsupervised representation learning, which can be roughly categorized into generative and discriminative methods. Generative methods usually take the form of auto-encoder~\cite{le2013building,kingma2013auto} or generative adversarial network~\cite{radford2015unsupervised}, and learn the feature embedding network by reconstructing the original high-dimensional inputs from the low-dimensional latent representations. However, generative methods perform pixel-level image reconstruction that is computationally expensive~\cite{grill2020bootstrap}. By contrast, discriminative methods learn representations by solving different pretext tasks, 
such as recognizing the rotation of the image 
~\cite{gidaris2018unsupervised}, 
solving jigsaw puzzles~\cite{noroozi2016unsupervised}, and contrastive prediction task~\cite{ye2019unsupervised}. Among them, contrastive prediction task-based solutions currently achieve state-of-the-art performance in unsupervised representation learning for a wide range of tasks, such as image classification~\cite{chen2020simple,dosovitskiy2014discriminative} and object detection~\cite{xie2021detco}. However, as discussed in Sections~\ref{sec:background} and~\ref{sec:evaluation}, conventional contrastive learning approaches focus on learning general representations that are more related to the appearance and the identity of the subjects, which leads to poor performance on gaze estimation. By contrast, \SystemName incorporates a set of techniques that encourage the learning of gaze-related features to improve gaze estimation performance.

There are also works on unsupervised gaze representation learning. Yu et al.~\cite{yu2020unsupervised} proposed a method to learn low dimensional gaze representations 
by jointly training a gaze redirection network~\cite{yu2019improving} and a gaze representation network. However, the training of the gaze redirection network requires well aligned eye images, setting constraints on their approach. More recently, by constructing eye-consistent and gaze-similar image pairs, Sun et al.~\cite{sun2021cross} proposed the cross-encoder to learn gaze representations from unlabeled eye images. These two solutions rely on generative models 
for representation learning, which requires the computationally expensive pixel-level image reconstruction. By contrast, \SystemName builds upon contrastive learning to avoid the 
reconstruction process.

\subsection{Learning in the Frequency Domain}

\SystemName is also related to existing efforts in using compressed frequency domain representations for image-based learning tasks~\cite{delac2009face,hafed2001face,ghosh2016deep,ehrlich2019deep,chamain2020improving,xu2020learning}. Delac et al.~\cite{delac2009face} investigated the feasibility of using the DCT and the discrete wavelet transform (DWT) coefficients in the JEPG and JEPG2000 compressed domain for face recognition. In the deep learning regime, Ghosh and Chellappa~\cite{ghosh2016deep} demonstrated that by leveraging the feature extraction capability of DCT and incorporating DCT operation on the feature maps generated by the convolutional layers, one can speed up the network convergence during training. Gueguen et al.~\cite{gueguen2018faster} suggested that when using the DCT coefficients as the inputs, the first fewer layers of CNN can be pruned to improve computational efficiency. Distinct from existing works, we exploit the spectral compaction property of DCT for aggressive input compression, and are the first to investigate unsupervised learning in the frequency domain for gaze estimation. 

\section{Challenges}
\label{sec:background}




\subsection{Challenges in Unsupervised Gaze Representation Learning}
\label{subsec:challengesInCGRL}

Motivated by recent advances in contrastive visual representation learning~\cite{chen2020simple,dosovitskiy2014discriminative,henaff2020data,he2020momentum}, we propose a contrastive learning (CL)-based framework for gaze representation learning. However, existing CL methods are designed for general purpose visual tasks, such as image classification~\cite{chen2020simple,dosovitskiy2014discriminative} and object detection~\cite{xie2021detco}, which cannot be used directly for gaze estimation. This is because, for general purpose tasks, existing solutions aim to identify a feature embedding network that 
ensures the feature representations of \textit{visually similar images}, i.e., images contain the same object instance or objects of the same category~\cite{wang2016contextual}, are close to each other in the low-dimensional embedding space~\cite{hadsell2006dimensionality}, while the feature representations of \textit{visually distinct images} are as separated as possible~\cite{ye2019unsupervised}. 
Unfortunately, for gaze estimation, \textit{facial images with the same gaze label can be visually distinct}, which makes the conventional contrastive learning methods ill-suited for gaze estimation. 

\begin{figure}[]
	\centering
	\includegraphics[width=8.5cm]{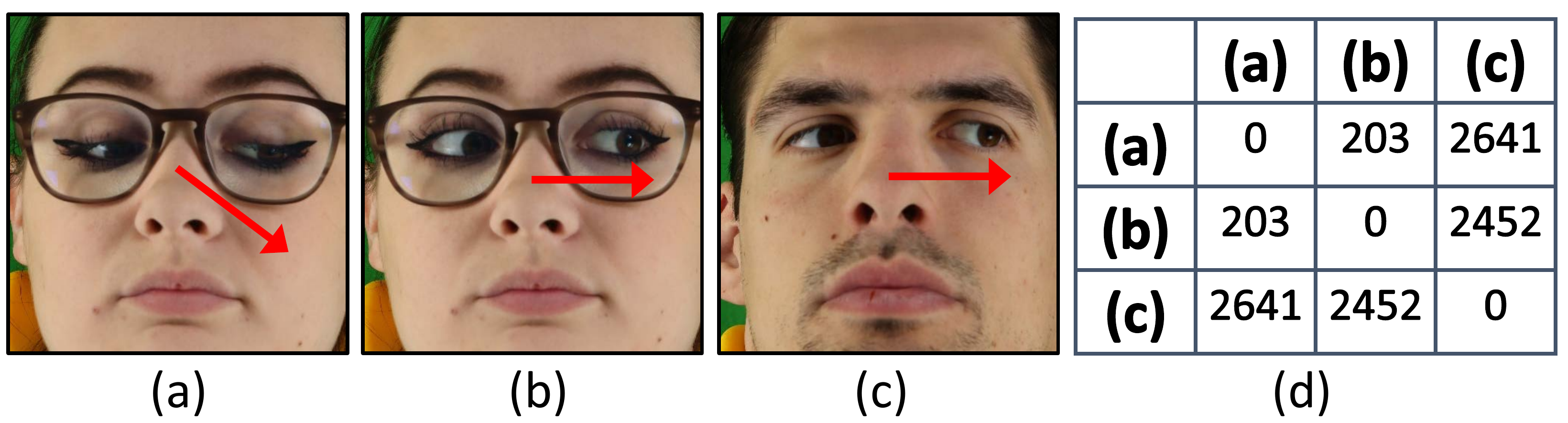}
	\caption{Examples of facial image labeled with gaze direction (red arrow): (a) and (b) are facial images of the same subject but have different gaze directions; in comparison, (b) and (c) are facial images of different subjects but have the same gaze direction; (d) shows the Euclidean distance between the learned representations of the three facial image pairs. 
	}
	\label{fig:gazeSementicDifference}
	\vspace{-0.15in}
\end{figure}

As an example, Figure~\ref{fig:gazeSementicDifference} shows three facial images labeled with the corresponding gaze direction. Figures~\ref{fig:gazeSementicDifference}(a) and (b) belong to the same subject, but have different gaze directions. In comparison, Figures~\ref{fig:gazeSementicDifference}(b) and (c) belong to different subjects, but have the same gaze direction. Conventional CL methods will treat the first two images as a visually similar pair, as they contain the same subject with highly similar appearance, while consider the latter two as a visually distinct pair, as they contain different subjects. As a result, the embedding network will learn representations that are related to subject identity, which are ill-suited for gaze estimation. 

To demonstrate, we adopt the widely used contrastive learning framework SimCLR~\cite{chen2020simple} to train a ResNet-18 based~\cite{he2016deep} embedding network for unsupervised representation learning. We use the ETH-XGaze~\cite{zhang2020eth} as the training dataset. After training, we take the three facial images shown in Figures~\ref{fig:gazeSementicDifference}(a)-(c) as the inputs of the embedding network to obtain the corresponding representations. 
Finally, we measure the Euclidean distance between the representations of the three images. The results are shown in Figure~\ref{fig:gazeSementicDifference}(d). 
The distance between image pair $(b, c)$ is 12 times larger than that of image pair $(a, b)$, even though $(b, c)$ have the same gaze direction. This indicates that the embedding network trained by SimCLR tends to push the representations of the \textit{visually similar} image pair $(a, b)$ close to each other in the embedding space, while separate the representations of the \textit{visually distinct} image pair $(b, c)$ apart. As shown in Section~\ref{subsec:gazeEstimationPerformance}, conventional contrastive learning will lead to 
high error in gaze estimation.
 


\subsection{Challenges in System Latency}
\label{subsec:challengeSystemLatency}

In addition to the data challenge, the associated high system latency also hinders the pervasive adoption of existing solutions. 

\textbf{Calibration latency.} The first type of latency is introduced by the {calibration} process of gaze estimation system. Subjects have large diversity in their eye shapes and inner eye structures~\cite{levin2011adler}, which introduces a non-negligible impact on the appearance of eyes and makes it difficult to build a ``one size fits all'' gaze estimation model. A widely adopted remedy for this problem is to perform subject-specific calibration before the system's first use~\cite{hansen2009eye,sugano2015self,lanata2015robust,linden2019learning}. Note that, even for the commercial high-end eye tracking systems, e.g., Tobii Pro~\cite{tobiiCalibration}, Pupil Lab~\cite{pupilLabCalibration}, and VIVE Pro Eye VR headset~\cite{vevoCalibration}, subject-specific calibration is also inevitable to ensure good gaze estimation accuracy. For deep learning-based systems, the calibration {requires} fine-tuning the pre-trained gaze estimation model by a small set of labeled gaze data collected from the targeted subject \textit{on the fly}. Moreover, given the fact that state-of-the-art gaze estimation systems adopt complex neural networks~\cite{cheng2021appearance}, e.g., ResNet-50~\cite{zhang2020eth} and VGG-16~\cite{zhang2017mpiigaze}, as the backbone network, the online calibration becomes a time-consuming process and affects the user's experience and acceptance of the system, especially when limited computational resource is available. 

\textbf{Inference latency.} Similar to the time demand in calibration, gaze estimation systems also have a {constrained time budget for inference}. Low inference latency 
is desirable for a wide range of applications, e.g., foveated rendering in virtual reality~\cite{konrad2020gaze,albert2017latency}, and eye tracking-based psychological research~\cite{mele2012gaze}. 

In Section~\ref{subsec:timeEfficiency}, our {latency evaluation} shows that a ResNet-18 based solution (which is 32-layers shallower than the ResNet-50 based design that is widely adopted by existing solutions~\cite{zhang2020eth}) 
takes 10.2 minutes for calibration, despite a high-end GeForce RTX 3080Ti GPU is used. By contrast, our proposed \SystemName can bring the latency down to 1.8 minutes, which is a {5.67 times} speedup. Similarly, for inference, \SystemName achieves at least 1.59 times speedup when comparing with the conventional method.




\section{System Design}
\label{sec:contrastiveLearning}





\subsection{Overview}
\label{subsec:overview}

\begin{figure}[]
	\centering
	\includegraphics[width=8.6cm]{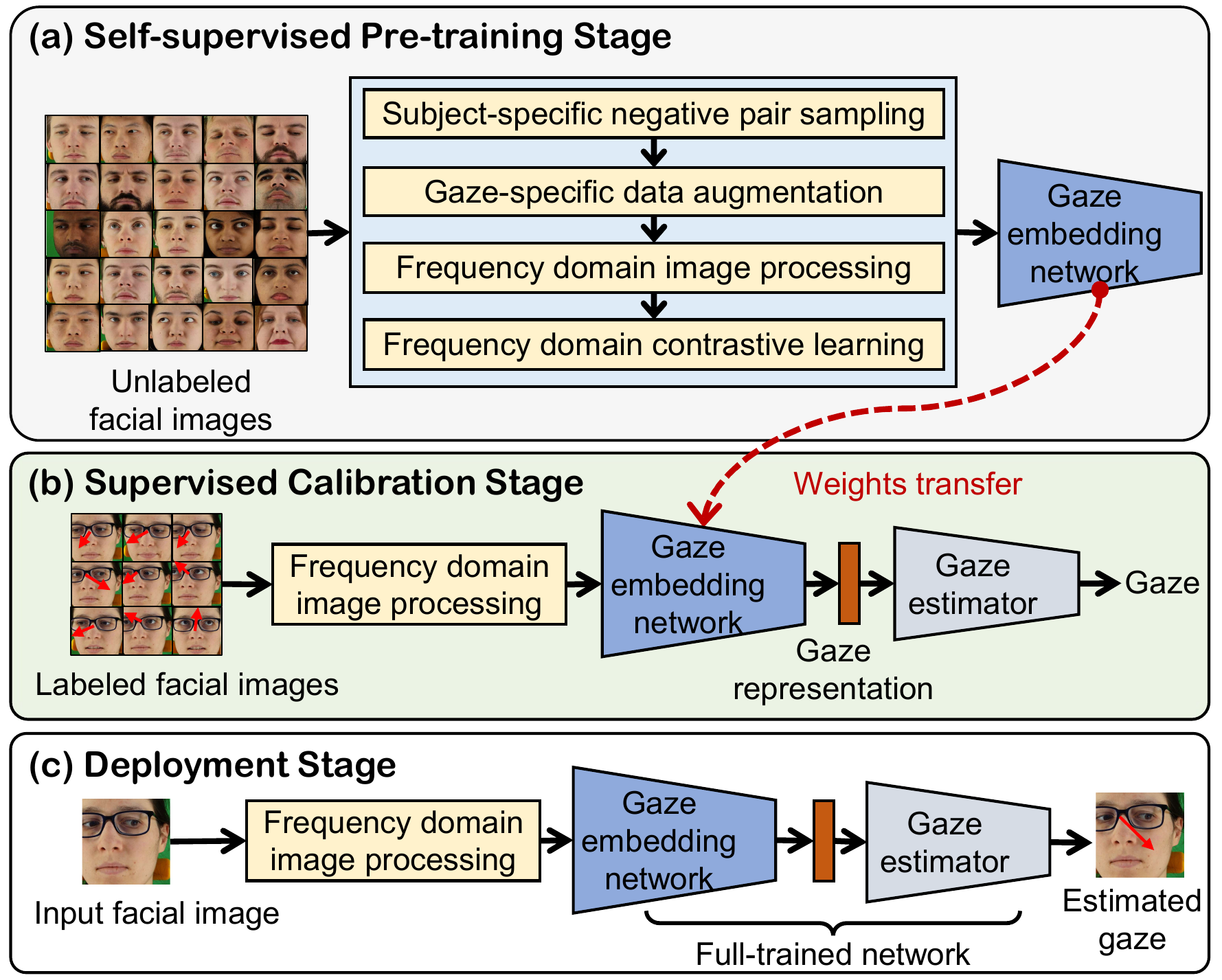}
	\caption{Overview of \SystemName which incorporates: (a) the {self-supervised pre-training stage}; (b) {supervised calibration stage}; and (c) {deployment stage}.}
	\label{fig:systemOverview}
	\vspace{-0.15in}
\end{figure}

In this work, we present \SystemName, a resource-efficient unsupervised learning framework for gaze estimation. At the core of \SystemName are the \textit{frequency domain gaze estimation} and the \textit{contrastive gaze representation learning}: the former dramatically reduces the latency in both system calibration and gaze estimation to ensure high time efficiency; while the latter overcomes the data labeling hurdle of existing supervised learning-based counterparts to ensure gaze {representation learning without gaze label. }
The overview design of \SystemName is shown in Figure~\ref{fig:systemOverview}, which includes three stages: the self-supervised pre-training stage, the supervised calibration stage, and the deployment stage. 

In the \textbf{self-supervised pre-training stage}, \SystemName takes unlabeled facial images as inputs to pre-train a \textit{gaze embedding network} (Section~\ref{subsec:gazeEmbeddingNetwork}) for unsupervised gaze representation learning. 
Specifically, we achieve this 
by a comprehensive set of techniques devised in this work, including: (1) the \textit{subject-specific negative pair sampling strategy} (Section~\ref{subsec:subjectSpecificNegativePairSampling}) that encourages the gaze embedding network to learn gaze-related features and abandon identity or appearance-related features; (2) the \textit{gaze-specific data augmentation} (Section~\ref{subsec:dataAugmentation}) that 
maintains the gaze-related semantic features in the augmented positive image pairs; (3) the \textit{frequency domain image processing module} (Section~\ref{subsec:DCT}) that selects the essential DCT coefficients of the original RGB facial images to reduce the computation latency in both {calibration and deployment stages}
; and (4) the \textit{frequency domain contrastive learning module} that leverages the DCT coefficients as the inputs to train the gaze embedding network. 

After the contrastive learning, the pre-trained gaze embedding network is transferred to the \textbf{supervised calibration stage}, and serves as a feature extractor for the downstream gaze estimation task. Specifically, taking a small amount of labeled facial images from the targeted subject as inputs, we first leverage the frequency domain image processing module to obtain the DCT coefficients of the original RGB images. Then, we fine-tune the pre-trained gaze embedding network and the gaze estimator for subject-specific gaze estimation. Note that, as mentioned in Section~\ref{subsec:challengeSystemLatency}, the subject-specific calibration is essential to ensure good gaze estimation accuracy, even for high-end eye tracking systems. Finally, in \textbf{the deployment stage}, the fine-tuned gaze embedding network and the gaze estimator are used for run-time gaze estimation. 

Below, we present the detailed design of \SystemName. We first introduce the pipeline of the frequency domain contrastive learning framework, followed by the designs of each individual components.

\subsection{Frequency Domain Contrastive Gaze Representation Learning}
\label{subsec:CL}



Inspired by recent advances in contrastive visual representation learning~\cite{chen2020simple,dosovitskiy2014discriminative,henaff2020data,he2020momentum}, we propose the frequency domain contrastive learning to train the gaze embedding network. The training is conducted by solving a set of contrastive prediction tasks. 
More specifically, the gaze embedding network learns the good representations for gaze estimation by maximizing the representation similarity between positive image pairs that have the same gaze direction, 
while minimizing the representation similarity between negative image pairs that have low gaze similarity. 
The pipeline of the proposed 
learning framework is shown in Figure~\ref{fig:frequencyCLOverview}, which consists of the following six major components:



\begin{figure}[]
	\centering
	\includegraphics[width=8.5cm]{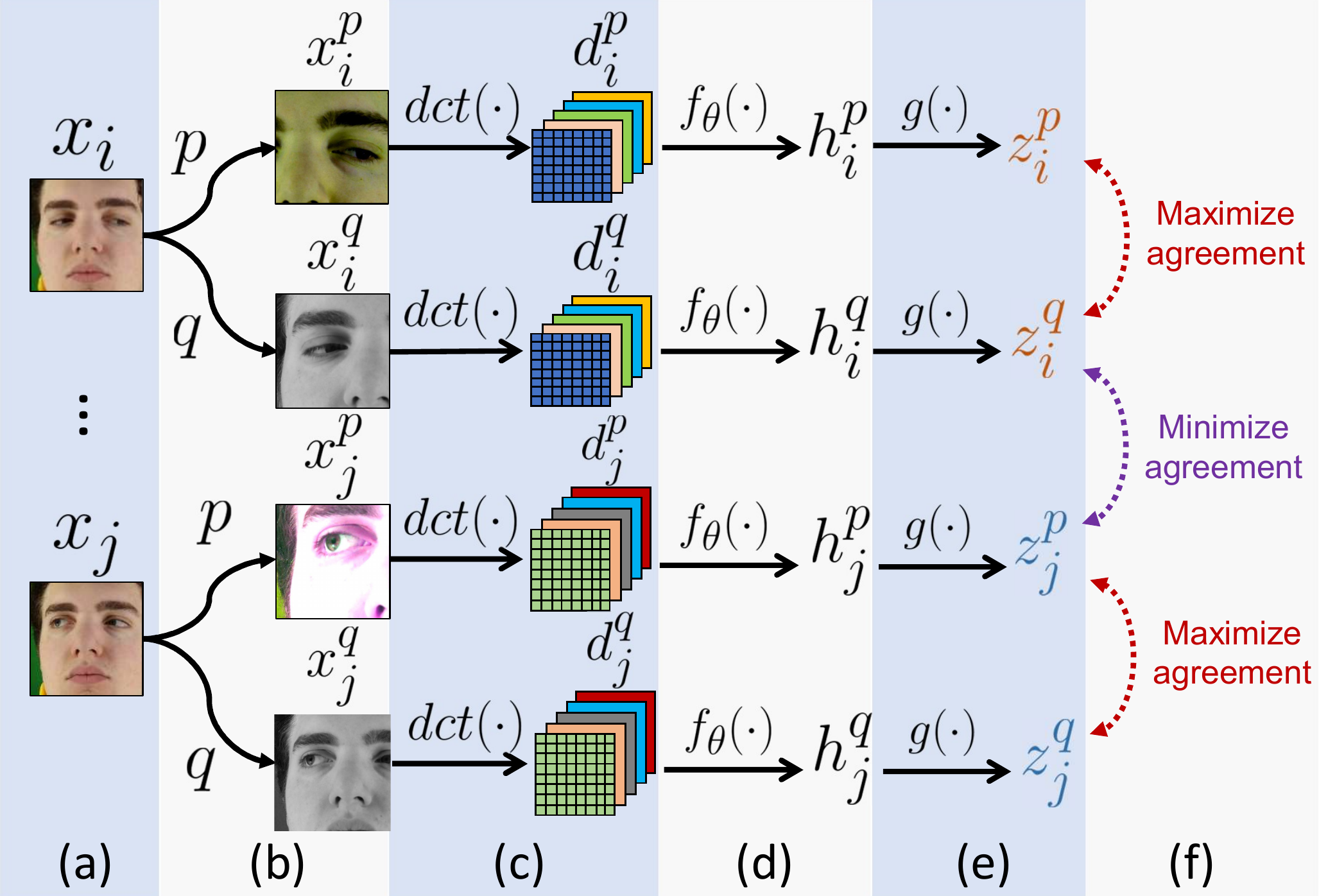}
	\caption{The pipeline of the proposed frequency domain contrastive gaze representation learning framework, which consists of: (a) subject-specific negative pair sampling; (b) stochastic gaze-specific data augmentation; (c) frequency domain image processing; (d) gaze embedding network; (e) projection head; and (f) contrastive prediction task.} 
	\label{fig:frequencyCLOverview}
	\vspace{-0.15in}
\end{figure}


\textbf{(a) Subject-specific negative pair sampling:} we first sample a minibatch of $N$ samples, $\{x_i\}_{i=1}^N$, from the unlabelled facial image set for contrastive learning. 
Instead of performing random image sampling, we introduce the \textit{subject-specific negative pair sampling strategy} (in Section~\ref{subsec:subjectSpecificNegativePairSampling}) that encourages the gaze embedding network to {focus on} gaze-related semantic features for better representation learning. As demonstrated in Section~\ref{subsubsec:dctCompare}, the proposed sampling strategy improves the gaze estimation accuracy by a large margin when comparing with the conventional method. 

\textbf{(b) Stochastic gaze-specific data augmentation:} we then apply the stochastic gaze-specific data augmentation to each image $x_i$ in the minibatch for two times. Specifically, 
we transform each $x_i \in \{x_i\}_{i=1}^N$ into two correlated views, denoted as $x_i^p$ and $x_i^q$, respectively. 
As shown in Figure~\ref{fig:frequencyCLOverview}, we consider images $x_i^p$ and $x_i^q$, which are augmented from the same facial image $x_i$, as a positive image pair. By contrast, $x_i^q$ and $x_j^p$, which are augmented from different image samples, are considered as a negative image pair. 
We introduce a novel gaze-specific data augmentation operator, i.e., \textit{gaze-cropping and resizing} (in Section~\ref{subsec:dataAugmentation}), to maintain the gaze-related semantic features for the augmented images. As demonstrated in Section~\ref{subsec:gazeEstimationPerformance}, our proposed method 
leads to better performance in gaze representation learning, and reduces the gaze estimation error by a large margin of 7$^\circ$ in angular error, when comparing with the conventional data augmentation operators~\cite{chen2020simple}.

\textbf{(c) Frequency domain image processing}: to reduce the computation latency in both calibration and gaze inference, we propose to use the DCT coefficients of the original RGB images as the inputs for representation learning. Specifically, we introduce the {frequency domain image processing}, $dct(\cdot)$, to transform the augmented image, $x_i^{t}$, from the RGB color space to the DCT frequency domain.
{The final output $d_i^t$ is the selected DCT coefficients matrix.} The detail of the frequency domain image processing is introduced in Section~\ref{subsec:DCT}. As shown in the latency evaluation in Section~\ref{subsec:timeEfficiency}, leveraging the DCT coefficients as inputs for representation learning can achieve up to 6.81 and 1.67 times speedup in system calibration and gaze inference, respectively.

\textbf{(d) Gaze embedding network}: then we train a {gaze embedding network}, $f_{\theta}(\cdot)$, to map the processed DCT coefficients matrix $d_i^t$ to the gaze representations in the latent space:
{
\begin{align} 
  h_i^t = f_{\theta}(d_i^{t}) \in \mathbb{R}^{D}, \; t \in \{p, q\} \; \textrm{and} \; i \in \{1,..,N\}, 
\end{align}}%
where $D$ is the dimension of the gaze representations. 
In Section~\ref{subsec:gazeEmbeddingNetwork}, we introduce the design of the gaze embedding network in detail.


\textbf{(e) Projection head:} we then use a nonlinear projection head, $g(\cdot)$, to further transform the gaze representations, $h_i^t$, to a lower dimensional latent space. The projection head is a multi-layer perceptron (MLP) with one hidden layer. The representations, $z_i^t$, after the nonlinear projection can be obtained by: 
{
\begin{align} 
  z_i^t = g(h_i^t), \; t \in \{p, q\} \; \textrm{and} \; i \in \{1,..,N\}.  
\end{align}}
The nonlinear projection head 
can encourage the embedding network $f_{\theta}(\cdot)$ to form and maintain features that are beneficial for the downstream task~\cite{chen2020simple,chen2020big}, which can further improve the quality of the learned representations $h_i^t$. 





\textbf{(f) Contrastive prediction task:} 
the gaze embedding network $f_{\theta}(\cdot)$ learns the good representations for gaze estimation by maximizing the representation similarity between positive image pairs that have {the same gaze direction}, while minimizing the representation similarity between negative image pairs that have different gaze directions. In another words, we aim to jointly train $f_{\theta}(\cdot)$ and { $g(\cdot)$ to pull }
the projected representations of the positive pairs, e.g., $z_i^p$ and $z_i^q$, closer together on the latent space; while pushes the representations of the negative pairs farther apart. 
This learning objective is achieved by solving the contrastive prediction task on the sampled minibatch $\{x_i\}_{i=1}^N$. Formally, {for two augmented sets $\{x_i^p\}_{i=1}^N$ and $\{x_i^q\}_{i=1}^N$ generated from $\{x_i\}_{i=1}^N$, the contrastive prediction loss function 
for a positive pair $(x_i^p, x_i^q)$ is defined as~\cite{wu2018unsupervised,chen2020simple}:}
{
\begin{align} 
\mathcal{L}_{i,p,q} = -\log{\frac{\exp \Big(\text{sim}(z_i^p,z_i^q)/\tau\Big)}{\sum_{k=1}^N{\exp\Big(\text{sim}(z_i^p,z_k^q)/\tau\Big)}}}, \; i \in \{1,..,N\},
\label{contrastiveLoss}	
\end{align}}%
where $\text{sim}(u,v)=u^\top v/\|u\|\|v\|$ is the cosine similarity of two feature vectors $u$ and $v$, and $\tau$ is a temperature parameter. 

The total loss is computed across all positive pairs in {$\{x_i^p\}_{i=1}^N$ and $\{x_i^q\}_{i=1}^N$}
for both $(x_i^p, x_i^q)$ and $(x_i^q, x_i^p)$, and can be obtained by:
{\begin{align} 
\mathcal{L} = \frac{1}{2N}\sum_{i=1}^N{\Big(\mathcal{L}_{i,p,q} + \mathcal{L}_{i,q,p}\Big)}.
\label{final loss}
\end{align}}%
Finally, 
$f_{\theta}(\cdot)$ and $g(\cdot)$ are jointly trained by minimizing 
$\mathcal{L}$. When the training is complete, the projection head will be discarded in the supervised calibration stage, and we use the gaze representations, $h_i^t$, as the input for the gaze {estimator, which is simply a MLP with one hidden layer.}

Below, we introduce the proposed subject-specific negative pair sampling strategy (Section~\ref{subsec:subjectSpecificNegativePairSampling}), stochastic gaze-specific data augmentation (Section~\ref{subsec:dataAugmentation}), frequency domain image processing (Section~\ref{subsec:DCT}), and the gaze embedding network (Section~\ref{subsec:gazeEmbeddingNetwork}) in detail.  

\subsection{Subject-specific Negative Pair Sampling}
\label{subsec:subjectSpecificNegativePairSampling}


\subsubsection{Rationale} Recall our discussion in Section~\ref{subsec:challengesInCGRL} that facial images with the same gaze direction can be visually distinct, while visually similar images can have completely different gaze directions (in Figure~\ref{fig:gazeSementicDifference}). Thus, to ensure good representation learning for gaze estimation, 
the gaze embedding network $f_{\theta}(\cdot)$ should 
learn representations that are related to the gaze, instead of capturing features that are related to the identity or the appearance of the subjects. As discussed in Section~\ref{subsec:CL}, $f_{\theta}(\cdot)$ learns the good representations by 
minimizing the contrastive loss defined in Equation~(\ref{contrastiveLoss}), which can be further extended as:
{\small
\begin{align}
\mathcal{L}_{i,p,q} = -\log{\frac{\exp \Big(\text{sim}(z_i^p,z_i^q)/\tau\Big)}{\exp \Big(\text{sim}(z_i^p,z_i^q)/\tau\Big) + \sum_{k\neq i}^N{\exp\Big(\text{sim}(z_i^p,z_k^q)/\tau\Big)}}}. 
\label{contrastiveLossRewrite}	
\end{align}}%
Clearly, minimizing $\mathcal{L}_{i,p,q}$ requires maximizing $\exp (\text{sim}(z_i^p,z_i^q)/\tau)$ and minimizing $\sum_{k\neq i}^N{\exp(\text{sim}(z_i^p,z_k^q)/\tau)}$. More specifically:
\begin{itemize}
[wide, labelwidth=!, labelindent=0pt]  
\item[\textbf{(1)}] maximizing the first term requires increasing the cosine similarity, $\text{sim}(z_i^p,z_i^q)$, {between the representations, $(z_i^p,z_i^q)$, obtained from the \textbf{positive pair}, $(x_i^p, x_i^q)$,} 
which helps $f_{\theta}(\cdot)$ to \textit{learn representations, $h_i^p$ and $h_i^q$, that are invariant to the data augmentations} that have been applied to 
$x_i$; 
\vspace{0.05in}
\item[\textbf{(2)}] minimizing $\sum_{k\neq i}^N{\exp(\text{sim}(z_i^p,z_k^q)/\tau)}$ requires reducing the cosine similarity between {the representations of the \textbf{negative pairs} $\{(x_i^p,x_k^q)\}_{k\neq i}$, } 
which enforces $f_{\theta}(\cdot)$ to \textit{learn representations that can differentiate $x_i$ from all the {other images in the minibatch}
}.
\end{itemize}

We can conclude that \textit{the selection of positive and negative pairs determines what representations the gaze embedding network can learn}. Thus, the strategy used in forming the minibatch, i.e., the sampling of the negative pairs, and the data augmentations that are used to generate the positive pairs, are crucial to ensure effective gaze representation learning. 

\subsubsection{Proposed method} 
Motivated by the above analysis, we propose the \emph{subject-specific negative pair sampling strategy} that encourages the gaze embedding network $f_{\theta}(\cdot)$ to learn gaze related features. More specifically, in each training iteration, all images in the minibatch $\{x_i\}_{i=1}^N$ are sampled from the same subject, such that the major visual difference among the negative pairs is caused by different gaze directions. Thus, by minimizing $\sum_{k\neq i}^N{\exp(\text{sim}(z_i^p,z_k^q)/\tau)}$, $f_{\theta}(\cdot)$ can learn representations that can differentiate images from the same subject with different gaze directions. In fact, the proposed sampling strategy implicitly uses the \textit{subject identity} as a prior knowledge during the training, and exploits the pairwise relations between the negative pairs, i.e., similar in appearance but different in gaze, to {force the gaze embedding network concentrate on gaze-specific visual features than features related to subject appearance and identity. }
By contrast, when constructing the minibatch, conventional contrastive learning algorithms~\cite{chen2020simple,chen2017sampling} randomly sample $N$ images from the dataset. Thus, the negative image pairs may contain different subjects, which makes the major visual difference among the negative pairs relate to subject appearance and identity. 

\subsection{Stochastic Gaze-specific Data Augmentation}
\label{subsec:dataAugmentation}


\subsubsection{Rationale} 
As discussed in Section~\ref{subsec:CL}, the gaze embedding network learns good representations for gaze estimation by maximizing the representation similarity between \textit{positive image pairs that have the same gaze direction}. In contrastive learning, positive image pairs are generated by applying different data augmentation operations {sampled from the stochastic data augmentation operators} on the same image. Many data {augmentation operators}
, e.g., cropping, rotation, cutout, and flip, have been proven to be useful for general purpose vision tasks, such as image classification~\cite{chen2020simple,dosovitskiy2014discriminative} and object detection~\cite{xie2021detco}. However, they are ill-suited for gaze representation learning, as they can \textit{mistakenly change or remove the gaze-related semantic features 
}. For instance, applying rotation and flip operations on the facial image will generate incorrect positive image pairs that have different gaze directions. Similarly, random cropping may reserve the mouth and the nose information of the subject, but remove the gaze-related ocular area features completely out from the {original images. }
As a result, conventional data augmentation {operators }
will add confusions to the contrastive learning, and lead to poor gaze estimation performance (in Section~\ref{subsec:gazeEstimationPerformance}). 
To resolve this problem, we propose \emph{gaze-cropping and resizing} as a novel data augmentation {operator} for gaze representation learning, which can maintain the gaze-related semantic features in the augmented positive image pairs. 

\subsubsection{Design} 
\begin{figure}[]
	\centering
	\includegraphics[width=8.4cm]{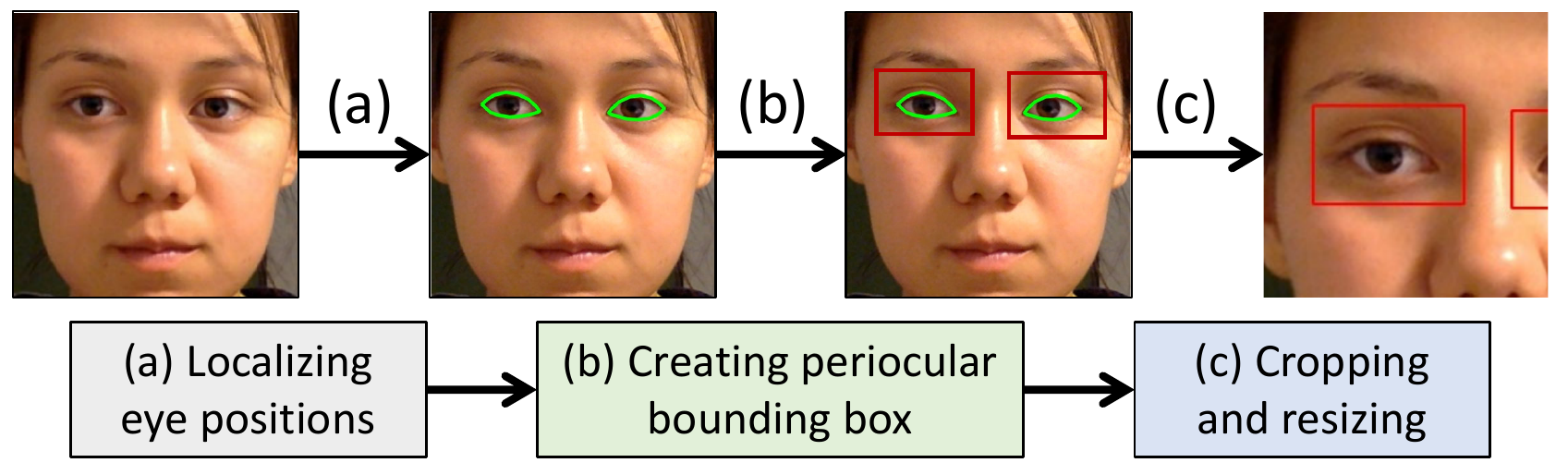}
	\caption{Pipeline of gaze-cropping and resizing: (a) localizing eye positions; (b) creating periocular bounding boxes; and (c) cropping and resizing.}
	\label{fig:gazeCropping}
	\vspace{-0.15in}
\end{figure}

The pipeline of the proposed data augmentation operator is shown in Figure~\ref{fig:gazeCropping}. First, we leverage the MediaPipe Face Mesh~\cite{google} to locate the eyes in the original facial image. The detected ocular areas are segmented in green ellipses in Figure~\ref{fig:gazeCropping}. Then, we create two periocular bounding boxes for the two eyes based on the estimated eye locations. As highlighted in red rectangles, the size of the periocular bounding box is larger than the detected eye such that the bounding box covers both the ocular and the periocular regions. After that, we randomly crop a patch from the facial image based on the locations of the bounding boxes. Specifically, we make sure that the cropped patch covers at least one of the periocular bounding boxes. Finally, we upsample the cropped image to the size of the original image. Since most of the gaze-related semantic features are contained in the ocular and the periocular areas~\cite{wood2015rendering,buhler2019content}, the proposed data augmentation operator ensures that the gaze-related semantic features are maintained in the resulting positive image pairs. 

Existing work in contrastive learning shows that the 
composition of multiple data augmentation {operators}
is crucial for learning good representations~\cite{chen2020simple}. Thus, when generating positive image pairs, we adopt \emph{color distortion} as the second data augmentation {operator} and apply it sequentially with the proposed gaze-cropping and resizing on the original image. While the gaze-cropping and resizing involves spatial/geometric transformation of the original image, the color distortion can change the appearance information of the original image, 
which makes the contrastive prediction tasks more difficult to solve and yields more effective representations~\cite{chen2020simple,kalantidis2020hard}.

\subsection{Frequency Domain Image Processing}
\label{subsec:DCT}





Below, we first investigate the impact of the input shape on the time complexity of the CNN, followed by the details of the proposed frequency domain image processing.

\subsubsection{Rationale} 
The time complexity of a {convolutional layer }
is {influenced }
by its FLOPs, memory consumption, and number of parameters~\cite{yao2018fastdeepiot}. {For a given CNN with a stack of convolutional layers}, the FLOPs of the $l$-th convolutional layer, denoted as $F_l(I)$, can be calculate by~\cite{he2015convolutional}:
\begin{equation} 
    F_l(I) = n_{l-1}\cdot s_l^2\cdot n_l \cdot m_l^2,
    \label{FLOPs}
\end{equation}
where $I$ is the original data input, $n_{l-1}$ is the number of input channels of the $l$-th convolutional layer, $s_l^2$ is the spatial size of the filter, i.e., the size of the first two dimensions, $n_l$ is the number of filters in the $l$-th convolutional layer, and $m_l^2$ is the spatial size of the output feature map. We also use $O_l(I)=n_l \cdot m_l^2$ to denote the output size of the $l$-th convolutional layer. 

Similarly, the memory consumption of the $l$-th convolutional layer, denoted as $M_l(I)$, can be calculated by~\cite{yao2018fastdeepiot}: 
\begin{equation} 
    M_l(I) = m_{l-1}^2\cdot n_{l-1}+m_{l}^2\cdot n_{l}+m_{l}^2\cdot s_l^2 \cdot  n_{l-1},
    \label{memory}
\end{equation} where $m_{l-1}^2$ is the spatial size of the input feature map.

As a toy example to motivate our design, we consider two network inputs 
\textit{with the same size but different shapes}: input $I_1$ has the shape of $224\times224\times3$, while input $I_2$ has the shape of $28\times28\times192$. {The spatial size of $I_1$ is 64 times of that of $I_2$, i.e., $(224\cdot 224)/(28 \cdot28) = 64$.} {We assume the first convolutional layer of the CNN has 64 filters and the filter spatial size is 3$\times$3. }
We also assume that the stride of the convolutional layer is one and padding is used. 

For the \textbf{first convolutional layer}, its output shape is $ 224\times224\times64$ and $28\times 28 \times 64$, for $I_1$ and $I_2$, respectively. Based on Equation~(\ref{FLOPs}), 
$I_1$ and $I_2$ lead to the same FLOPs for the first convolutional layer:
\begin{equation} 
    F_1(I_1) = 3\cdot3^2\cdot64\cdot224^2 = F_1(I_2) = 192\cdot3^2\cdot64\cdot28^2.
\end{equation}
Based on Equation~(\ref{memory}), we can also calculate the corresponding memory consumption as:
\begin{equation} 
    M_1(I_1) = 224^2\cdot3+224^2\cdot64+224^2\cdot3^2\cdot3 = 4,716,544.
\end{equation}
\begin{equation} 
    M_1(I_2) = 28^2\cdot192+28^2\cdot64+28^2\cdot3^2\cdot192 = 1,555,456.
\end{equation}
Thus, for the first convolutional layer, $I_1$ and $I_2$ result in the same FLOPs, but $M_1(I_1)$ is more than three times of $M_1(I_2)$.

For the \textbf{$l$-th convolutional layer when $l\geq2$}, we can further rewrite Equations~(\ref{FLOPs}) and (\ref{memory}) as:
\begin{equation} 
    F_l(I) = n_{l-1}\cdot s_l^2\cdot O_{l}(I),
    \label{FLOPs2}
\end{equation}
\begin{equation} 
    M_l(I) =O_{l-1}(I)+O_{l}(I)\left(1 + s_l^2\cdot (\frac{n_{l-1}}{n_l})\right).
    \label{memory2}
\end{equation}
Moreover, from the second convolutional layer, 
the values of $s_l$, $n_{l-1}$, and $n_l$ are independent from the original network input $I$. Thus, as shown in Equations~(\ref{FLOPs2}) and (\ref{memory2}), we can consider both $F_l(I)$ and $M_l(I)$ as functions of $O_{l}(I)$. Finally, given $O_1(I_1)$ is 64 times of $O_1(I_2)$ in our example, we have $F_l(I_1) \approx 64 \cdot F_l(I_2)$ and $M_l(I_1) \approx 64 \cdot M_l(I_2)$, for any convolutional layer with $l\geq2$. Note that the magic number 64 comes from the {ratio} of the spatial size of the inputs. 
Based on the above analysis, we can draw a conclusion that \textit{{reducing the spatial size of the CNN input can dramatically reduce its FLOPs and memory consumption, and thus reduces the time complexity.}} In addition, if we can also reduce the number of channels of input, e.g., from $28\times 28 \times 192$ to $28 \times 28 \times 12$, both FLOPs and memory consumption can be further reduced. 

\subsubsection{Design}
Apparently, simply reshaping and compressing the original RGB input will destroy the perceptual 
information it contains. To resolve this challenge, we leverage the discrete cosine transform (DCT) to convert the inputs from the RGB domain to the frequency domain, and leverage the frequency domain DCT coefficients as the inputs for gaze estimation. This enables us to reshape the original input without losing the essential perceptual information. Moreover, the \textit{spectral compaction}~\cite{rao2014discrete} property of the DCT transformation tells us that most of the critical content-defining information of the image is concentrated in the low-end of the frequency spectrum, whereas signals in the high-frequency end are mostly trivial and are associated with noise~\cite{wallace1991jpeg}. Therefore, we can leverage this property to significantly compact the essential perceptual information in the facial image by a few DCT coefficients in the low-frequency domain. In fact, this is a well-known property that has been widely exploited by data compression techniques, such as the JPEG compression standard~\cite{pennebaker1992jpeg}, where the quantization process rounds off most of the high frequency components during image compression~\cite{chen1984scene}. Our method is orthogonal to existing efforts in model compression~\cite{cheng2018model} and data offloading~\cite{yao2020deep}, and can be used in parallel to further reduce the system latency. 

\begin{figure}[]
 	\centering
	\includegraphics[width=8.3cm]{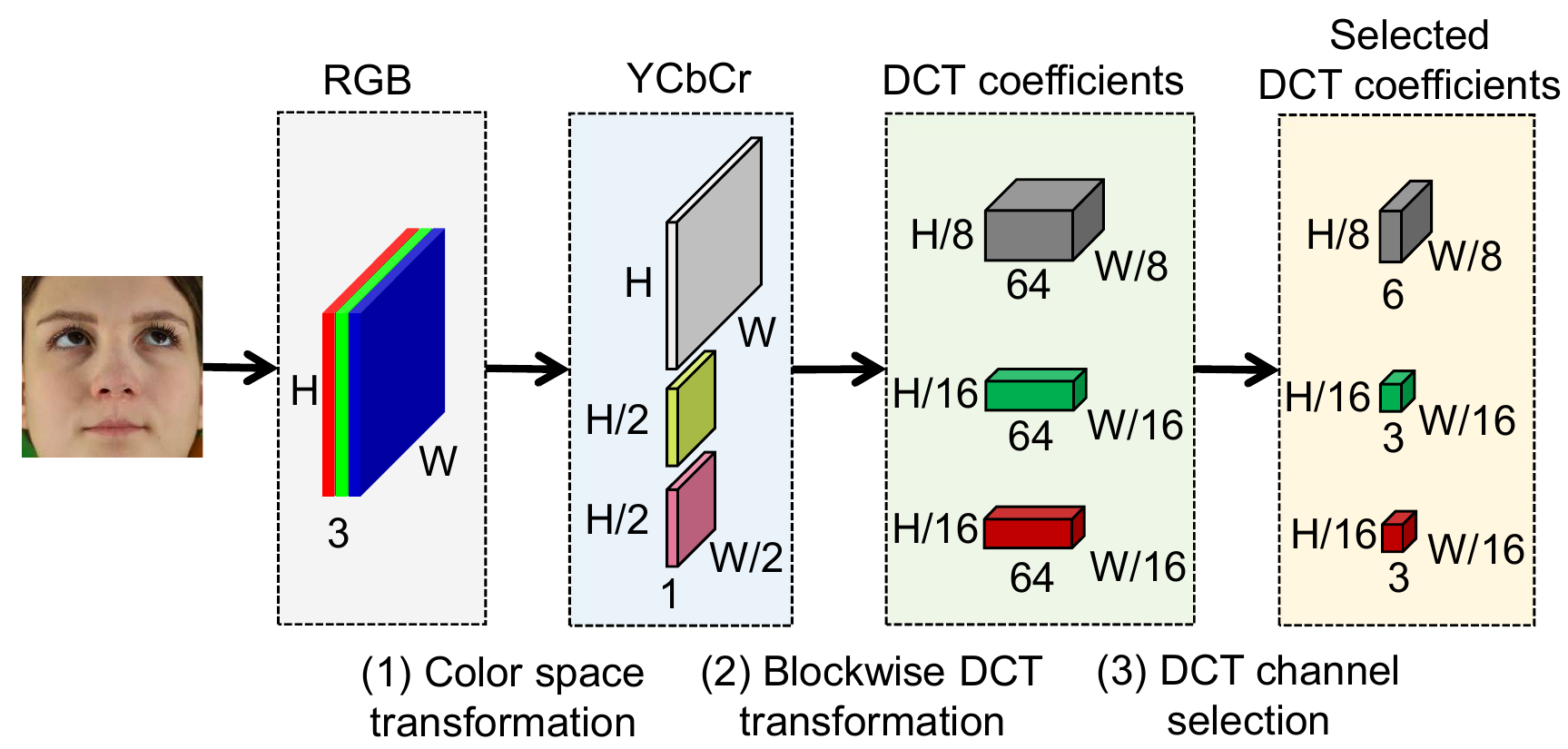}
	\caption{The pipeline of the proposed frequency domain image processing. The input RGB facial image is (1) first converted to the YCbCr color space with the chroma components downsampled. (2) Then, we apply the blockwise DCT transformation on the three components to obtain the corresponding DCT coefficients matrix. (3) Finally, we perform the DCT channel selection to retain only the essential DCT coefficients as the input for gaze estimation. 
	}
	\label{fig:dctProcessing}
	\vspace{-0.15in}
\end{figure}

The pipeline of the proposed frequency domain image processing is shown in Figure~\ref{fig:dctProcessing}. It takes an RGB facial image with the shape $W\times H \times 3$ as the input, where $W$ and $H$ denotes the width and height, 
respectively. First, we apply \textbf{color space transformation} to convert the input image from the RGB color space to the YCbCr color space, which results in three components: one \textit{luma component} (Y), representing the brightness, and two \textit{chroma components} (Cb and Cr), representing the chrominance of the image. As shown in Figure~\ref{fig:dctProcessing}, the spatial resolution of the Cb and Cr components is reduced by a factor of two. This is because human visual system is more sensitive to fine-grained brightness details, but less susceptible to the hue and color details of an image. Thus, the two chroma components can be compressed~\cite{pennebaker1992jpeg,zhang2021intelligent}. 

After the color space transformation, each of the Y, Cb, and Cr components is partitioned into multiple $8 \times 8$ rectangular nonoverlapping blocks. Then, we apply the \textbf{blockwise discrete cosine transform} (DCT) 
to obtain the frequency domain representations, i.e., the DCT coefficients. Specifically, given a block $\mathbf{B}\in\mathbb{R}^{8\times 8}$, 
the corresponding DCT coefficients are denoted by matrix $\mathcal{B}\in\mathbb{R}^{8\times 8}$, whose element $\mathcal{B}_{i,j}$ is obtained by~\cite{ahmed1974discrete}: 
{\small
\begin{align}
\mathcal{B}_{i,j} = s(i) s(j) \sum_{n=0}^{7}\sum_{m=0}^{7}\mathbf{B}_{n,m}\cos\Big[\frac{\pi}{8}\Big(n+\frac{1}{2}\Big)i\Big]\cos\Big[\frac{\pi}{8}\Big(m+\frac{1}{2}\Big)j\Big],
\end{align}}%
where $\mathbf{B}_{n,m}$ is the pixel value at coordinate $(n,m)$; integers ${0\leq i \leq 7}$ and ${0\leq j \leq 7}$ represent the horizontal and the vertical frequency, respectively; $s(i)$ is a normalization factor to make the transformation orthonormal, for which $s(i)=(\frac{1}{8})^{0.5}$ if $i=0$ and $s(i)=0.5$ 
otherwise. 
In sum, each block $\mathbf{B}$ is represented in the frequency domain by a weighted combination of 64 orthogonal sinusoids, where $\mathcal{B}_{i,j}$ is DCT coefficient that indicates the spectral energy. After the DCT transformation, we scan each of the $8\times8$ DCT coefficients matrices in a \textit{zigzag} order starting from the top-left corner and subsequently convert it to a $1\times64$ vector. As shown in Figure~\ref{fig:dctProcessing}, the outputs of the DCT transformation are three coefficients matrices, with the shape of $\frac{W}{8} \times \frac{H}{8} \times 64$, $\frac{W}{16} \times \frac{H}{16} \times 64$, and $\frac{W}{16} \times \frac{H}{16} \times 64$, for the Y, Cb, and Cr components, respectively. 

Finally, taking advantage of the spectral compaction property of DCT, we perform \textbf{DCT channel selection} to further compress the DCT coefficients matrix and retain a small subset of it as the data input for the gaze estimation. As shown in Figure~\ref{fig:dctProcessing}, instead of maintaining all the 64 channels, we only keep the lowest six, three, and three channels for the Y, Cb, and Cr components, respectively, and prune the other high-frequency channels. 
Taking an RGB image with the size of $224 \times 224 \times 3$ as the input, the frequency domain image processing outputs the selected DCT coefficients matrix of Y, Cb, and Cr, with the shape of $28 \times 28 \times 6$, $14\times 14\times 3$ and $14\times 14\times 3$, respectively. The selected DCT coefficients matrices are then fed into the gaze embedding network for representation learning.

\subsection{Gaze Embedding Network}
\label{subsec:gazeEmbeddingNetwork}

\begin{figure}[]
 	\centering
	\includegraphics[width=8.cm]{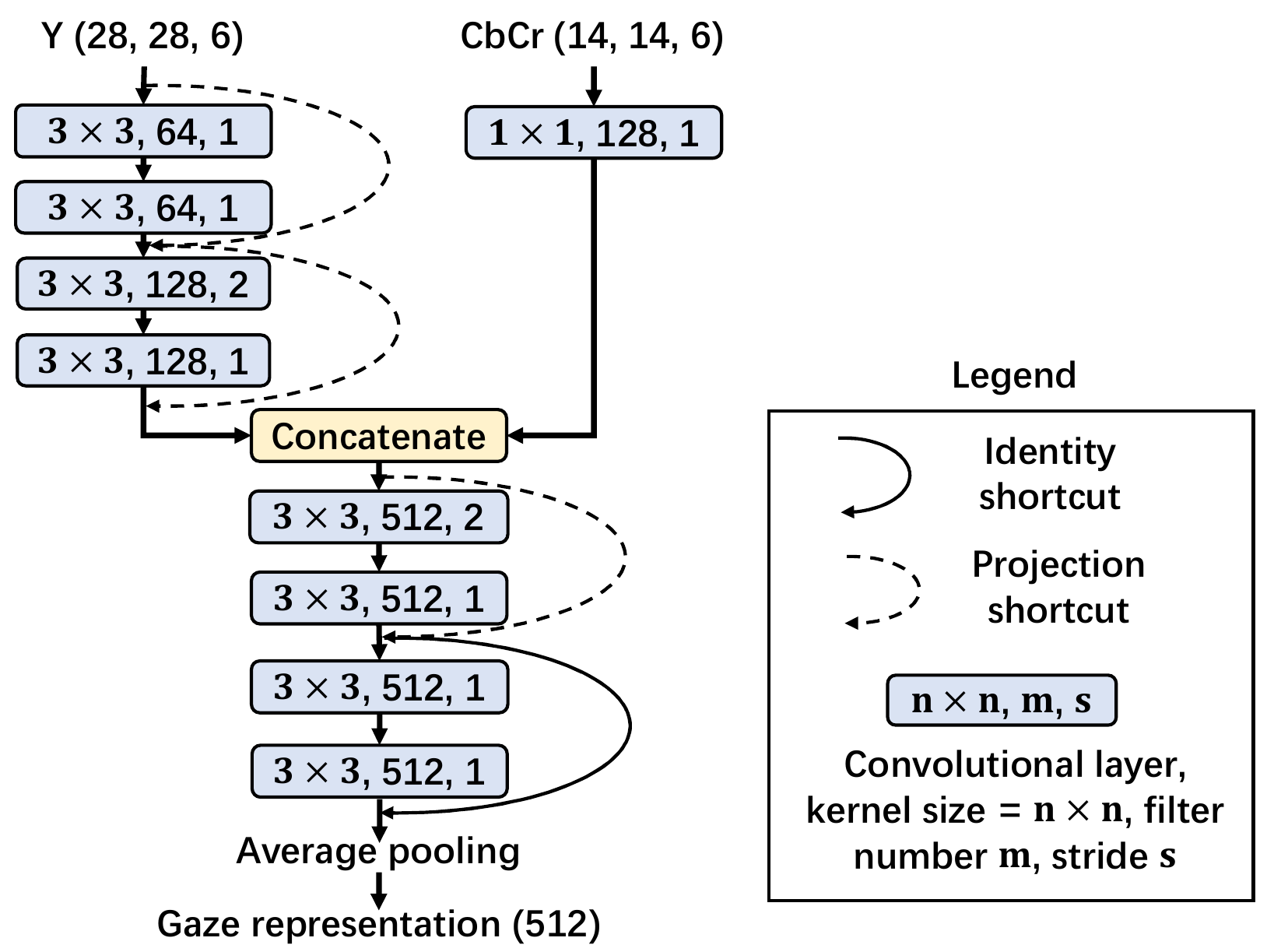}
	\caption{The architecture of the gaze embedding network. 
	}
	\label{fig:cnnArch}
	\vspace{-0.15in}
\end{figure}

Figure~\ref{fig:cnnArch} shows the architecture of the gaze embedding network, which follows a ResNet-18~\cite{he2016deep} based design. 
Since the DCT coefficients matrix of the Y component has a larger spatial size than that of the Cb/Cr components, we cannot feed the three DCT coefficients matrices directly to a conventional CNN. A naive solution is to downsample the input of Y component or upsample the input of the Cb/Cr components to align the spatial sizes. However, it will lead to information loss or a higher time complexity. 
Thus, we have to design the CNN network carefully so that it can take the three matrices with different sizes as inputs.
Specifically, we adopt the late concatenate
~\cite{gueguen2018faster} 
, which has approved to ensure better learning performance. As depicted in Figure~\ref{fig:cnnArch}, the DCT coefficients matrix of the Y component is first passed through two residual blocks consist of four convolutional layers, whose output shape is $14 \times 14 \times 128$. 
In parallel, the DCT coefficients matrices of Cb/Cr are first concatenated, followed by fed into a single convolutional layer, for which the output shape is $14 \times 14 \times 128$. Then, the outputs of two parallel paths are concatenated as they have the same spatial size. After that, the joint representations are fed into two residual blocks which correspond to block conv5 in the original ResNet-18 design. The final output of the gaze embedding network is the gaze representations, which is a vector of 512 features. {In our design, we apply batch normalization (BN)~\cite{ioffe2015batch} right after each convolution and before activation, and also adopt BN after the input layer to ensure better performance.} Note that the architecture shown in Figure~\ref{fig:cnnArch} contains eight convolutional layers in the deepest path, which is eight layers shallower than the original ResNet-18. We adopt the current design after balancing the gaze estimation accuracy and system latency. We provide a detailed investigation on this trade-off in Section~\ref{subsubsc:impactOfModelSize}.

\section{Evaluation}
\label{sec:evaluation}


\subsection{Datasets}
We consider two public gaze estimation datasets 
in our evaluation.

\begin{itemize}
[wide, labelwidth=!, labelindent=0pt]  
\item \textbf{ETH-XGaze}~\cite{zhang2020eth} is the state-of-the-art appearance-based gaze estimation dataset, which originally consists of one training and two testing sets. The images in the training set are released with gaze labels, whereas the labels for the testing sets are not available, i.e., testing results can only be evaluated via the project leaderboard~\cite{ETHXGazeLEADERBOARD}. Thus, we only take the original training set of ETH-XGaze for our evaluation, which consists of 80 subjects with diverse genders, ages, and ethnicity backgrounds. Some of the subjects wore contact lenses or eye glasses during the data collection. We consider the application scenario of using smartphone front-camera and web-camera for gaze estimation, and select images with the front-facing head pose. We further split the selected data into two parts: a \textit{pre-training set} comprised of 70 subjects, and a \textit{calibration set} consists of the remaining 10 subjects. Each subject has 380 to 600 images. In total, {the pre-training set has 36,731 images and the calibration set has 4,718 images}. The resolution of image is 448 $\times$ 448. 



\item \textbf{MPIIFaceGaze}~\cite{zhang2017mpiigaze} is collected from 15 subjects. Each subject has 3,000 images with diverse head poses and illumination conditions. We take the images collected from the first ten subjects to form the pre-training set, and that from the remaining five subjects to form the calibration set, which results in a size of 30,000 and 15,000 images for the pre-training and calibration set, respectively.
\end{itemize}



\noindent
\textbf{Data preparation.} For both datasets, the \textit{pre-training set} is used to pre-train the gaze embedding network, which is conducted in either a supervised or unsupervised way, depending on the representation learning method considered. For each subject in the \textit{calibration set}, the subject's data is further divided into three parts: (1) a \textit{fine-tuning set} for fine-tuning the pre-trained gaze embedding network and the gaze estimator in a subject-dependent manner; (2) a \textit{validation set} for network validation; and (3) a \textit{testing set} to evaluate the gaze estimation performance after the fine-tuning. Specifically, for each subject in the calibration set (ten subjects for the ETH-XGaze and five subjects for the MPIIFaceGaze), we randomly sample 75 images to form the fine-tuning set, 25 images to form the validation set, and use the remaining images 
to form the testing set. For each subject, we have 280 to 500 testing images for the ETH-XGaze and 2,900 testing images for the MPIIFaceGaze. In Section~\ref{subsec:personalizationPerformance}, we study how the fine-tuning set size affects the gaze estimation performance. 



\subsection{Methodology}
\label{subsec:methodology}

\subsubsection{Methods in comparison}
We compare the gaze estimation performance of different supervised and self-supervised methods. 

\begin{itemize}
[wide, labelwidth=!, labelindent=0pt]  

\item \textbf{\SystemName}: the proposed self-supervised method. 
Specifically, we first pre-train the gaze embedding network by following the frequency domain contrastive gaze representation learning introduced in Section~\ref{sec:contrastiveLearning}. Then, we fine-tune the gaze embedding network and the randomly initialized gaze estimator using the 
images in the fine-tuning set. 

\item \textbf{\SystemName-Mix}: a variant of \SystemName. Differently, during the self-supervised learning stage, images in the minibatch are randomly sampled from the whole pre-training set, without using the \textit{subject-specific negative pair sampling strategy} introduced in Section~\ref{subsec:subjectSpecificNegativePairSampling}. 

\end{itemize}
We also consider three baselines that take RGB images as the inputs:

\begin{itemize}
[wide, labelwidth=!, labelindent=0pt]  

\item \textbf{RGB-Supervised}: 
we leverage the pre-training set to pre-train the gaze embedding network and the gaze estimator in a supervised manner, i.e., \textit{37K labeled images for ETH-XGaze and 30K labeled images for MPIIFaceGaze}. We further fine-tune them during the calibration stage. 

\item \textbf{RGB-Random}: we randomly initialize the parameters of the gaze embedding network and the gaze estimator without pre-training. We only fine-tune them during the calibration stage. 

\item \textbf{RGB-SimCLR}: we leverage the state-of-the-art contrastive learning framework, i.e., SimCLR~\cite{chen2020simple}, to pre-train the gaze embedding network. Specifically, we adopt the data augmentation {operators} introduced in SimCLR, i.e., random cropping followed by image resizing and random color distortion, and apply them sequentially on the RGB images to generate positive and negative image pairs. Moreover, in self-supervised learning stage, images in the minibatch are randomly sampled from the whole pre-training set. The gaze estimator is randomly initialized. 
Both networks are then fine-tuned during the calibration stage.

\end{itemize}

Lastly, we consider \textbf{DCT-Supervised} and \textbf{DCT-SimCLR}. 
Specifically, {DCT-Supervised} leverages the DCT coefficients matrices generated by the frequency domain signal processing as the inputs for both pre-training and calibration. For {DCT-SimCLR}, we obtain the DCT coefficients of the RGB images that are augmented by the data augmentation methods in SimCLR. 
Different from \SystemName, {DCT-SimCLR} adopts neither the subject-specific negative pair sampling strategy nor the stochastic gaze-specific data augmentation. 

\subsubsection{Implementation}
\label{subsubsec:implementation}
We implement all the methods with Python and TensorFlow 2.0. For RGB-based methods, the gaze embedding network follows the design of ResNet-18 \cite{he2016deep} (without the dense layers). For DCT-based methods, the gaze embedding network follows the structure shown in Figure~\ref{fig:cnnArch}. We use the Adam optimizer~\cite{kingma2014adam} during the training. In the pre-training stage, all methods use the same batch size of 128. The learning rate for \SystemName, \SystemName-Mix, RGB-SimCLR, and DCT-SimCLR is 0.01; the learning rate for RGB-Supervised and DCT-Supervised is 0.001. For all methods, the resolution of the input images is $224\times224$. In the calibration stage, the learning rate for RGB-Supervised, RGB-SimCLR, and RGB-Random is 0.00025 with a decay of 0.0005. The learning rate for \SystemName, \SystemName-Mix, DCT-SimCLR is 0.002. The learning rate for DCT-Supervised is 0.0005 with a decay of 0.0005. Note that, for the four DCT-based methods, 
we use images with the resolution of $448\times448$ to fine-tune the network, which improves estimation accuracy. We do not apply this to the three RGB-based methods, 
as it leads to high calibration latency, i.e., 10 to 21 minutes as measured in Section~\ref{subsubsec:calibrationTime}, which makes it an impractical design choice. 
For fair comparison, all methods use the same pre-training set for either unsupervised or supervised pre-training. In the calibration stage, we randomly sample the fine-tuning and the validation sets for six times, and all methods use the same sets of data for fine-tuning and testing. We report the averaged result as the final accuracy.

\subsection{Performance in Gaze Estimation} 
\label{subsec:gazeEstimationPerformance}

\subsubsection{Performance metrics} 
\begin{figure}[]
 	\centering
	\includegraphics[width=7cm]{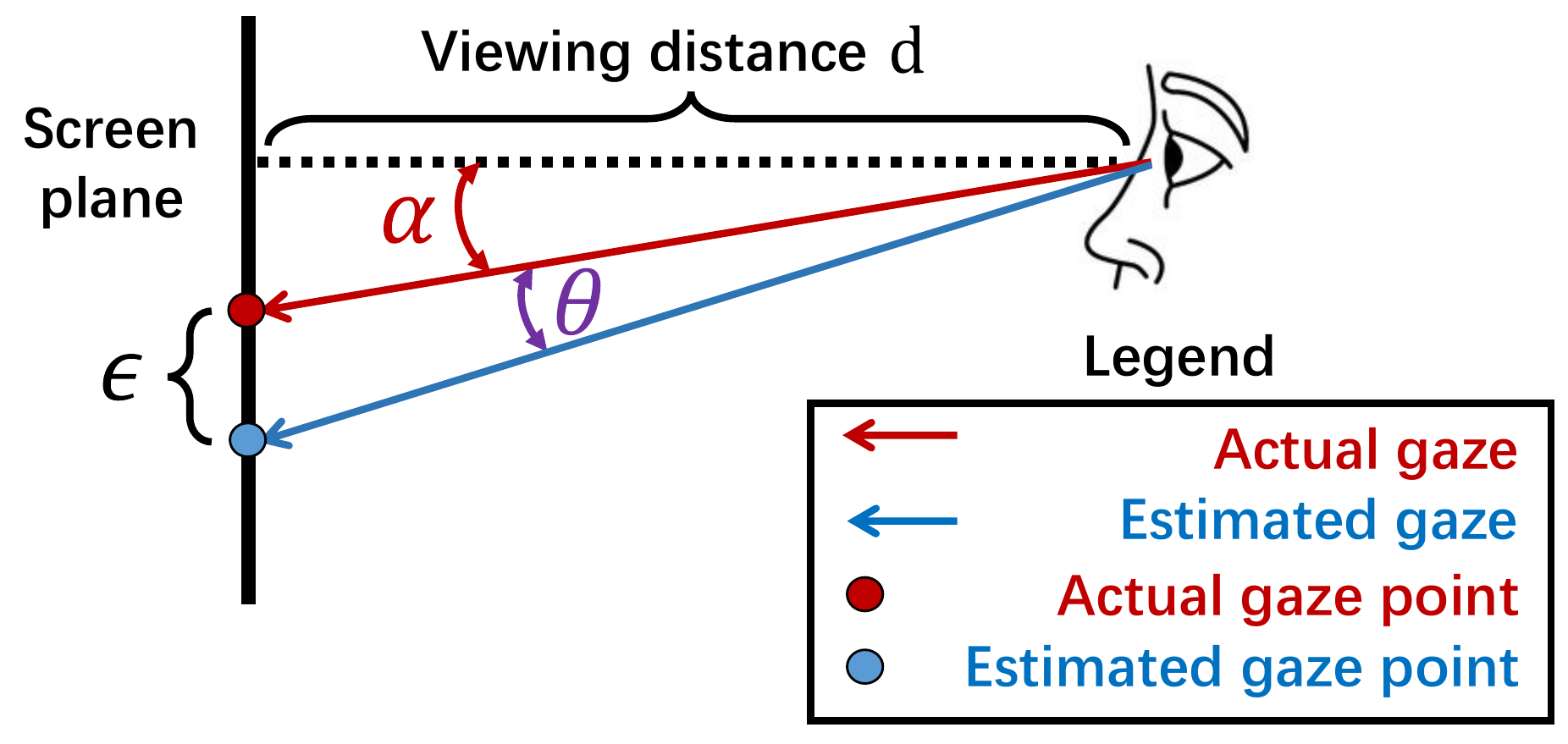}
	\caption{Illustration of the performance metrics used for gaze estimation: (1) the angular error, $\theta$, represents the angle (in degree) between the estimated and the actual gazes; the distance error, $\epsilon$, represents the euclidean distance (in cm) between the estimated and the actual gaze points on the screen plane, given a viewing distance $d$.
	}
	\label{fig:gazeErrorMetric}
	\vspace{-0.1in}
\end{figure}

As shown in Figure~\ref{fig:gazeErrorMetric}, we use the {angular error} $\theta$ (in degree) and the {distance error} $\epsilon$ (in cm) as the performance metrics. Specifically, $\theta$ represents the angle between the estimated and actual gaze vectors; while $\epsilon$ measures the euclidean distance between the estimated and actual gaze points on the screen plane, given the viewing distance $d$. 

\subsubsection{Comparison with RGB-based methods}
\begin{figure}[]
	\centering
	\subfigure[ETH-XGaze]{\includegraphics[scale=0.51]{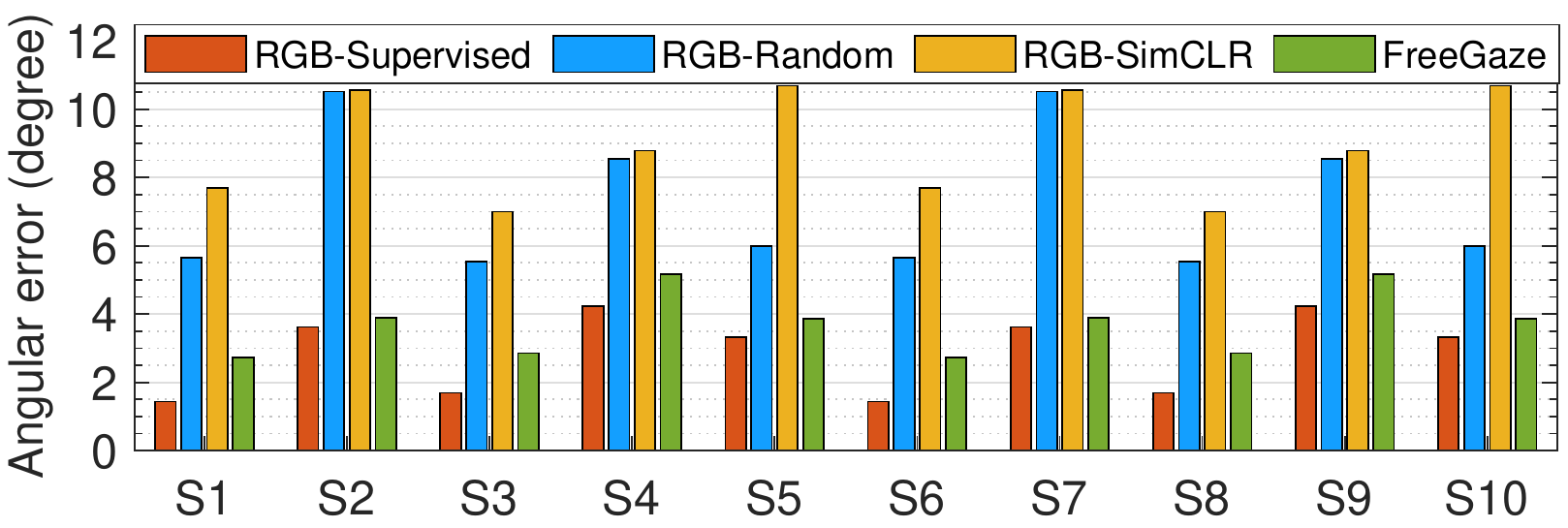}}\\[-1.5ex]
	\subfigure[MPIIFaceGaze]{\includegraphics[scale=0.51]{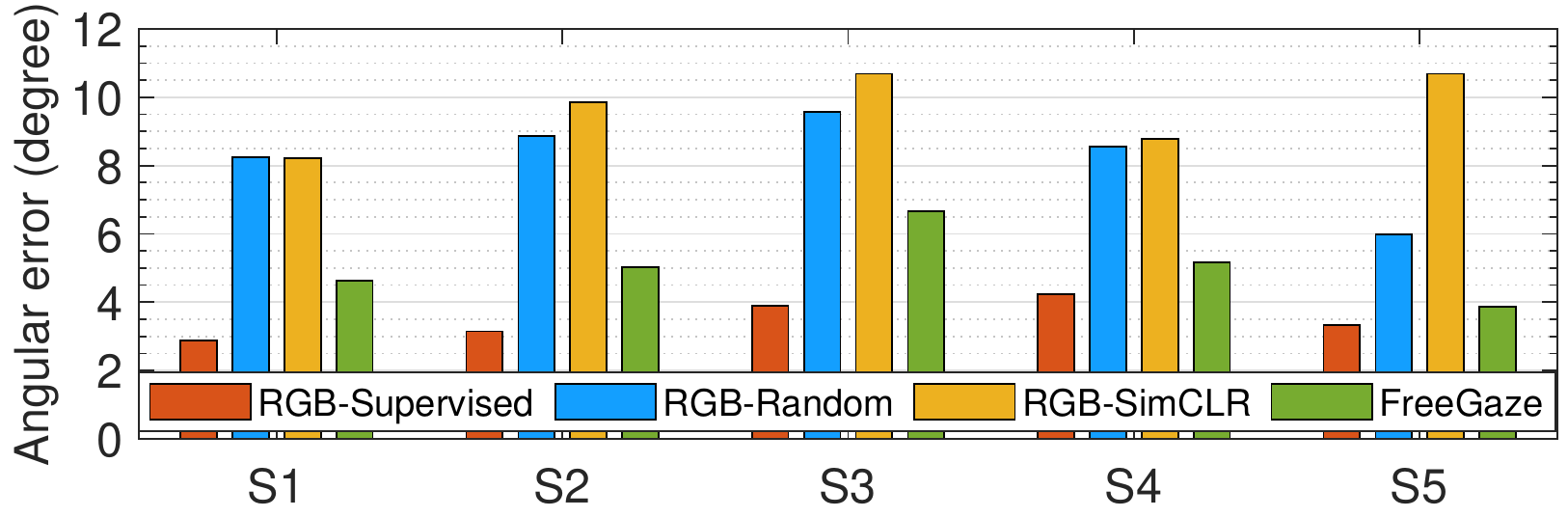}} 
	\caption{Angular error of the examined methods 
	on (a) ETH-XGaze and (b) MPIIFaceGaze. \SystemName achieves similar performance with RGB-Supervised, and outperforms RGB-Random and RGB-SimCLR by a large margin in all cases.}	
	\label{fig:angleError}
	\vspace{-0.1in}
\end{figure}

We compare the proposed \textit{\SystemName} with the RGB-based methods, i.e., \textit{RGB-Supervised}, \textit{RGB-Random}, and \textit{RGB-SimCLR}. 
Figure~\ref{fig:angleError} shows the angular error 
on each subject in the testing set. The averaged angular errors are given in {Table~\ref{tab:angleError}}.  
We make the following observations.

First, {RGB-Supervised}, which leverages 37K and 30K labeled RGB images to train the gaze embedding network and the gaze estimator, achieves the lowest angular error in all examined cases. By contrast, without the help from the supervised pre-training, the angular error of {RGB-Random} is significantly higher.

\begin{table}[]
    \caption{Averaged angular errors (mean$\pm$std) 
    on ETH-XGaze and MPIIFaceGaze.}
    \label{tab:angleError}
    \resizebox{2.9in}{!}{
    \begin{tabular}{c|c|c|}
    \cline{2-3}
    \multicolumn{1}{l|}{}                         & \textbf{ETH-XGaze} & \textbf{MPIIFaceGaze} \\ \hline
    \multicolumn{1}{|c|}{\textbf{RGB-Supervised}} & 1.81$^\circ\pm$0.75$^\circ$        & 3.50$^\circ\pm$0.56$^\circ$           \\ \hline
    \multicolumn{1}{|c|}{\textbf{RGB-Random}}     & 7.67$^\circ\pm$2.30$^\circ$        & 8.24$^\circ\pm$1.35$^\circ$           \\ \hline
    \multicolumn{1}{|c|}{\textbf{RGB-SimCLR}}     & 8.38$^\circ\pm$1.63$^\circ$        & 9.65$^\circ\pm$1.12$^\circ$           \\ \hline
    \multicolumn{1}{|c|}{\textbf{FreeGaze}}       & 2.95$^\circ\pm$0.51$^\circ$        & 5.07$^\circ\pm$1.03$^\circ$           \\ \hline
    \end{tabular}
    }
    \vspace{-0.15in}
\end{table}

Second, 
{RGB-SimCLR} has the highest angular error. This indicates that conventional SimCLR cannot be used directly for gaze estimation, and the gaze embedding network has poor representation learning performance when training by SimCLR. There are two reasons for this result. First, SimCLR performs random image cropping when generating the positive image pairs~\cite{chen2020simple}, which potentially removes the gaze-related features, 
and thus, adding confusions to the gaze embedding network during the representation learning. Second, when forming the minibatch, SimCLR randomly samples images from the whole pre-training set to generate negative pairs, and considers images from different subjects as negative pairs. However, as discussed in Section~\ref{subsec:challengesInCGRL} and demonstrated in Figure~\ref{fig:gazeSementicDifference},
contrastive learning ensures the representations of \textit{visually similar images}, i.e., images of the same subject, are close to each other in the latent space~\cite{hadsell2006dimensionality}, while the representations of \textit{visually distinct images} are as separated as possible~\cite{ye2019unsupervised}. Thus, when images of different subjects are considered as negative pairs, the gaze embedding network 
tends to learn representations that are useful for subject classification, instead of gaze estimation. These two factors make SimCLR ineffective for the aimed purpose.

Lastly, the proposed {\SystemName} achieves similar performance with {RGB-Supervised}, and out-performs both {RGB-Random} and {RGB-SimCLR} by a large margin. The performance gap in angular error between {\SystemName} and {RGB-Supervised} is 1.14$^\circ$ and 1.57$^\circ$ on ETH-XGaze and MPIIFaceGaze, respectively.

\begin{figure}[]
	\centering
	\subfigure[ETH-XGaze ($d=30cm$)]{\includegraphics[scale=0.44]{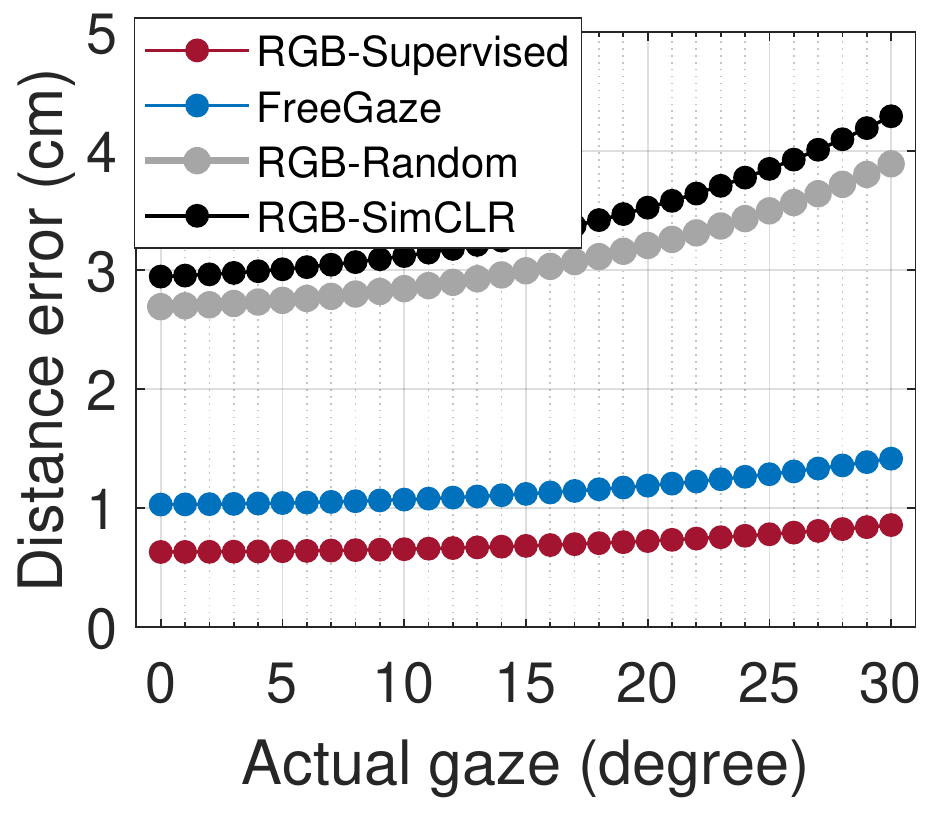}} 
	\subfigure[ETH-XGaze ($d=50cm$)]{\includegraphics[scale=0.44]{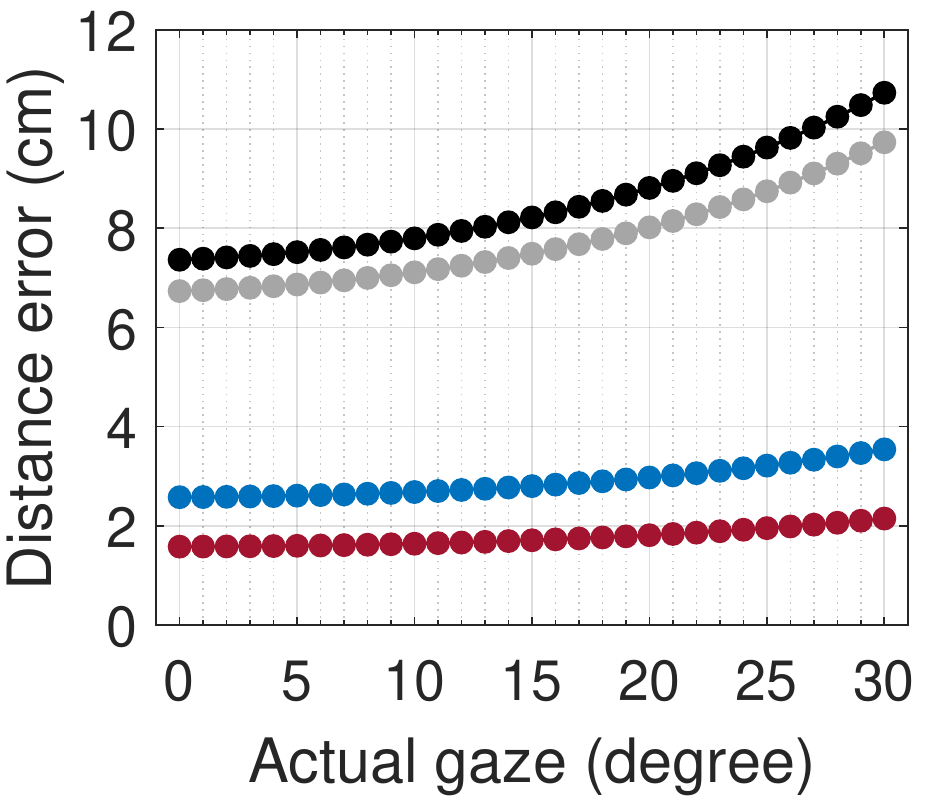}} \\[-2ex]
	\subfigure[MPIIFaceGaze ($d=30cm$)]{\includegraphics[scale=0.44]{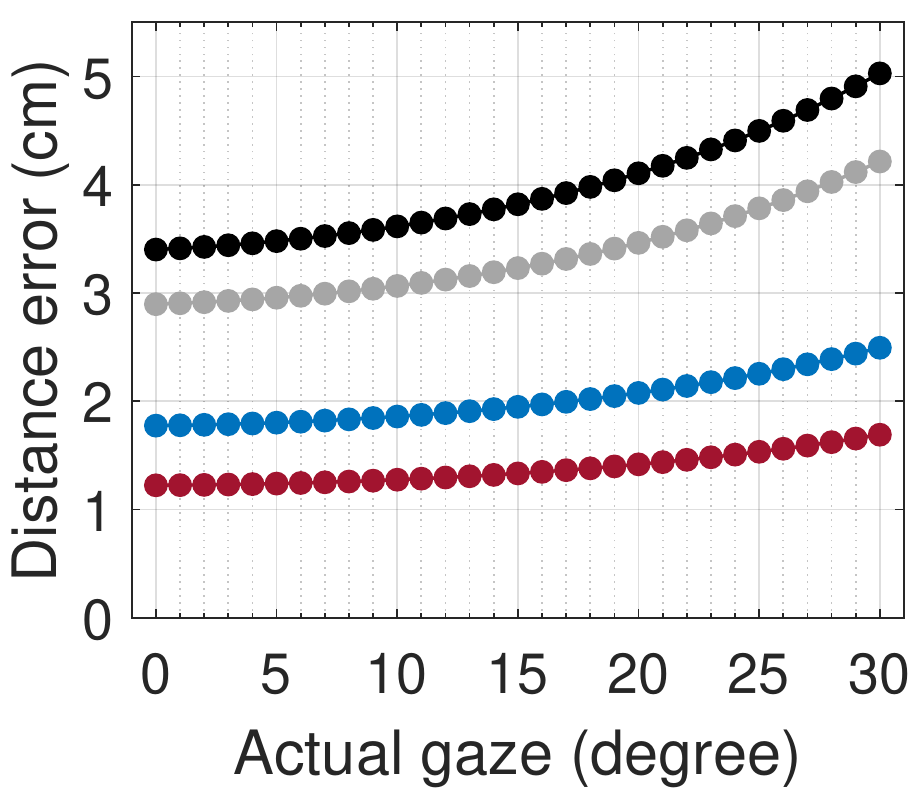}}
	\subfigure[MPIIFaceGaze ($d=50cm$)]{\includegraphics[scale=0.44]{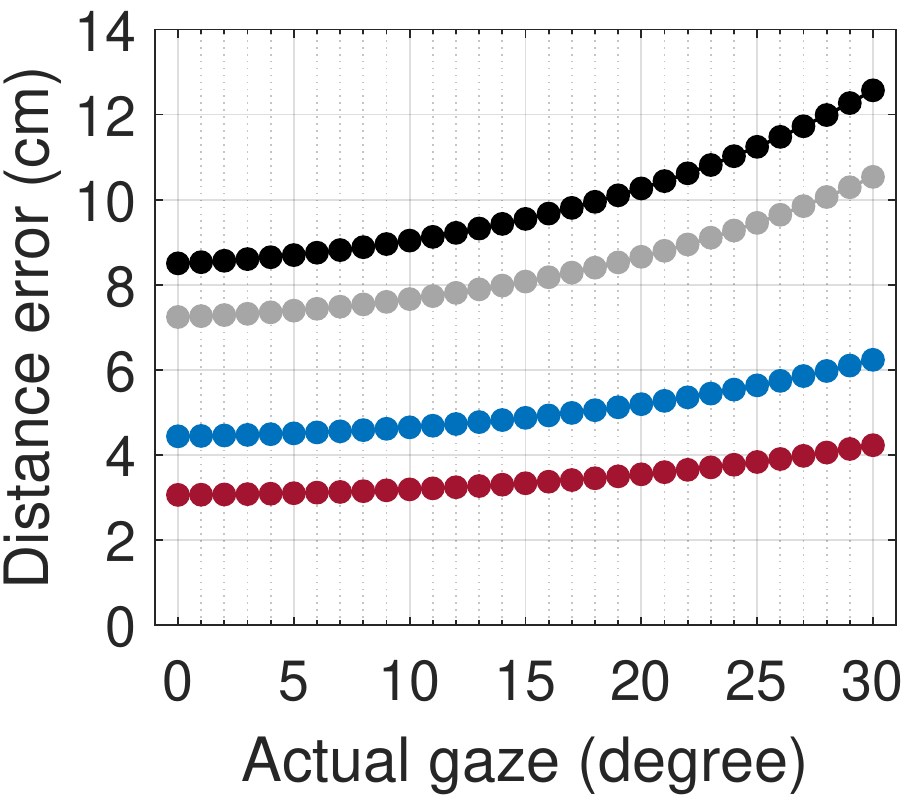}}\\[-1.5ex]
	\caption{Distance error $\epsilon$ 
	as a function of the viewing distance $d$ and the actual gaze angle $\alpha$.}
	\label{fig:distanceError}
	\vspace{-0.15in}
\end{figure}

\textbf{Performance in different application scenarios.} As shown in Figure~\ref{fig:gazeErrorMetric}, we also use the distance error $\epsilon$ as the performance metric. This allows us to quantify the gaze estimation performance with the consideration of typical application scenarios, e.g., eye tracking using front-facing camera on smartphone and desktop computer. 
Given the angular error $\theta$, we can calculate 
$\epsilon$ by:
{
\begin{align} \small
\epsilon = d \times \big(\tan(\alpha+\theta) - \tan(\alpha)\big),
\end{align}}%
where $d$ is the viewing distance and $\alpha$ is the actual gaze angle. Then, taking the averaged angular errors reported in Table~\ref{tab:angleError}, we calculate the corresponding distance error $\epsilon$ as a function of $d$ and $\alpha$. Specifically, we set $d$ equals 30cm and 50cm, respectively, to represent the typical viewing distance when a subject is using smartphone~\cite{boccardo2021viewing,long2017viewing} and desktop computer~\cite{rempel2007effects}, respectively. The actual gaze angle $\alpha$ changes from 0$^\circ$ to 30$^\circ$, which is the range of the standard viewing angle when interacting with digital visual displays~\cite{johnson1997hci,greeson1996international}. The averaged distance errors of the four methods are shown in Figure~\ref{fig:distanceError}. {\SystemName} outperforms {RGB-Random} and {RGB-SimCLR} by a large margin, and achieves acceptable performance when comparing with {RGB-Supervised}. Specifically, in the smartphone viewing scenario, i.e., $d=30$cm, the averaged performance gap (in distance error) between {\SystemName} and {RGB-Supervised} is only 0.45cm and 0.64cm, on ETH-XGaze and MPIIFaceGaze, respectively. In the desktop viewing scenario, i.e., $d=50$cm, the averaged performance gap increases slightly to 1.13cm and 1.59cm, on ETH-XGaze and MPIIFaceGaze, respectively.

\subsubsection{Comparison with DCT-based methods}
\label{subsubsec:dctCompare}
\begin{figure}[]
	\centering
	\subfigure[ETH-XGaze]{\includegraphics[scale=0.51]{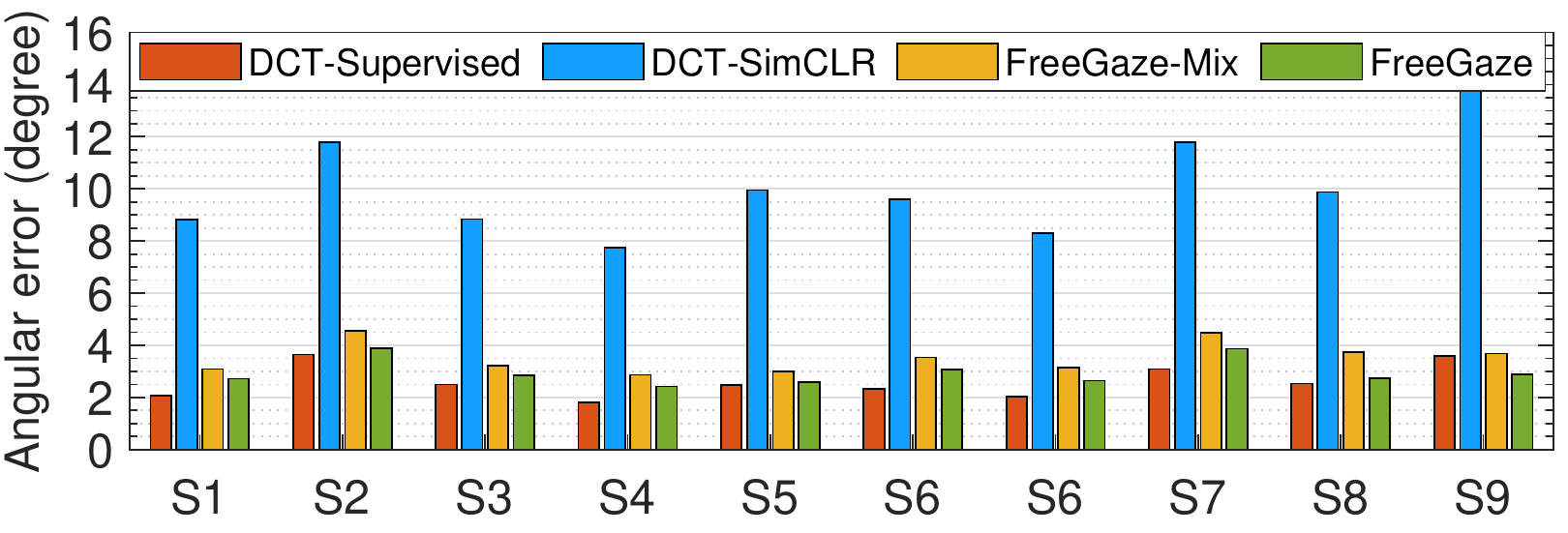}}\\[-1.5ex]
	\subfigure[MPIIFaceGaze]{\includegraphics[scale=0.51]{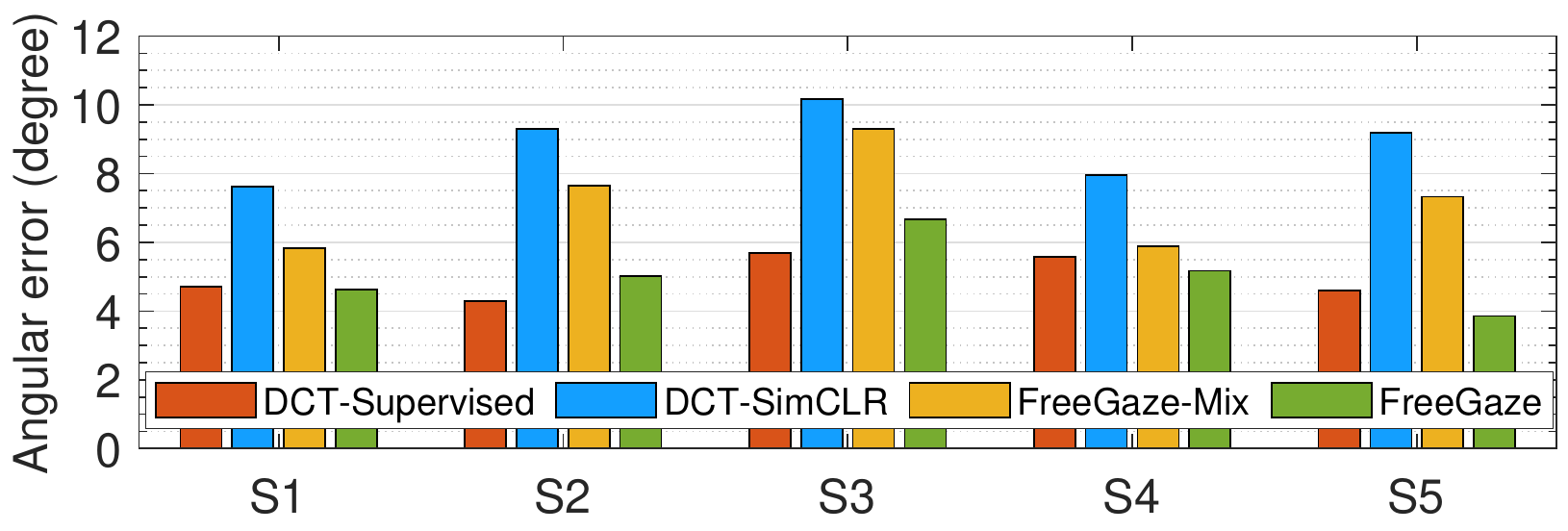}} 
	\caption{Angular error of the DCT-based methods 
	on (a) ETH-XGaze and (b) MPIIFaceGaze dataset. 
	}	
	\label{fig:angleErrorDCT}
	\vspace{-0.15in}
\end{figure}

We also compare the gaze estimation performance of the four DCT-based methods. The results 
are shown in Figure~\ref{fig:angleErrorDCT}.

Overall, on ETH-XGaze the angular error averaged on the ten subjects is 2.61$^\circ\pm$0.64$^\circ$, 10.11$^\circ\pm$2.00$^\circ$, 3.53$^\circ\pm$0.59$^\circ$, and 2.95$^\circ\pm$0.51$^\circ$, for DCT-Supervised, DCT-SimCLR, \SystemName-Mix, and \SystemName, respectively. Similarly, on MPIIFaceGaze the angular error averaged on the five subjects is 4.97$^\circ\pm$0.62$^\circ$, 8.84$^\circ\pm$1.04$^\circ$, 7.20$^\circ\pm$1.44$^\circ$, and 5.07$^\circ\pm$1.03$^\circ$, for the four methods, respectively. We make the following additional observations.

{DCT-Supervised} achieves the lowest angular error, as it leverages labeled DCT coefficients for supervised training. The proposed {\SystemName} achieves the lowest angular error amongst the three unsupervised methods. Specifically, {\SystemName} outperforms {\SystemName-Mix} by 
0.56$^\circ$ and 2.14$^\circ$ on average, for ETH-XGaze and MPIIFaceGaze, respectively. This improvement comes from the adoption of \textit{the subject-specific negative pair sampling strategy} introduced in Section~\ref{subsec:subjectSpecificNegativePairSampling}. By contrast, {\SystemName-Mix} leverages the conventional minibatch sampling strategy introduced in SimCLR, which treats images from different subjects as negative pairs. Nevertheless, \SystemName-Mix still outperforms DCT-SimCLR by a large margin: 
6.58$^\circ$ and 1.64$^\circ$ on average for the two datasets, respectively. This result further indicates that the random image cropping used by SimCLR adds confusions to the gaze embedding network during the representation learning, and leads to poor estimation performance. By contrast, using of \textit{the gaze-specific data augmentation} (introduced in Section~\ref{subsec:dataAugmentation}), both \SystemName and \SystemName-Mix can maintain the gaze-related features and ensure good representation learning capability. Lastly, the performance gap between \SystemName and DCT-Supervised is 
0.36$^\circ$ and 0.09$^\circ$ on average for the two datasets, which demonstrates the effectiveness of the proposed unsupervised learning framework for learning gaze representations in the frequency domain.

\subsubsection{Impact of fine-tuning set size}
\label{subsec:personalizationPerformance}
\begin{figure}[]
	\centering
	\subfigure[ETH-XGaze]{\includegraphics[scale=0.48]{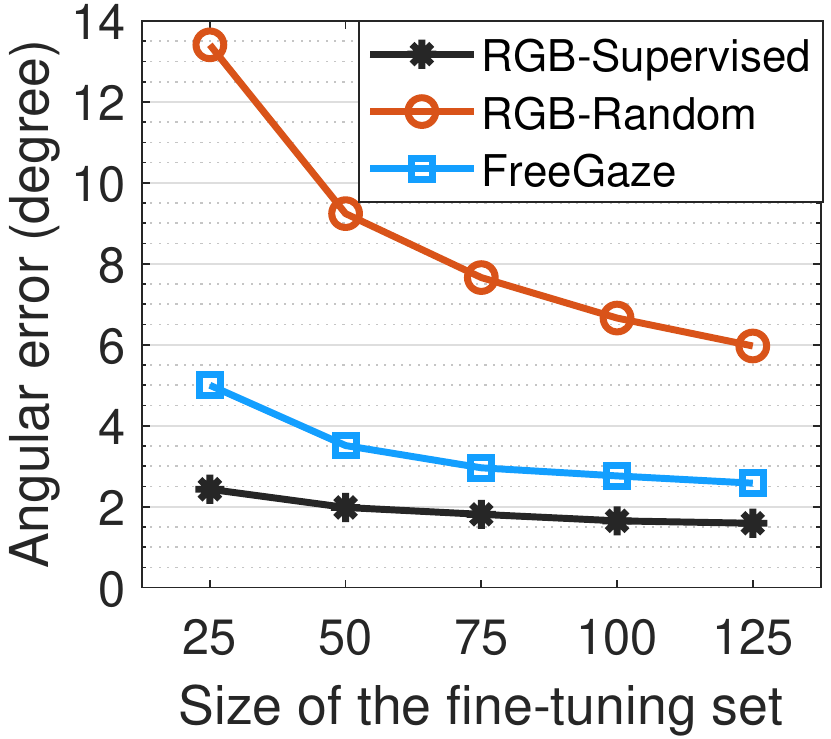}} 
	\subfigure[MPIIFaceGaze]{\includegraphics[scale=0.48]{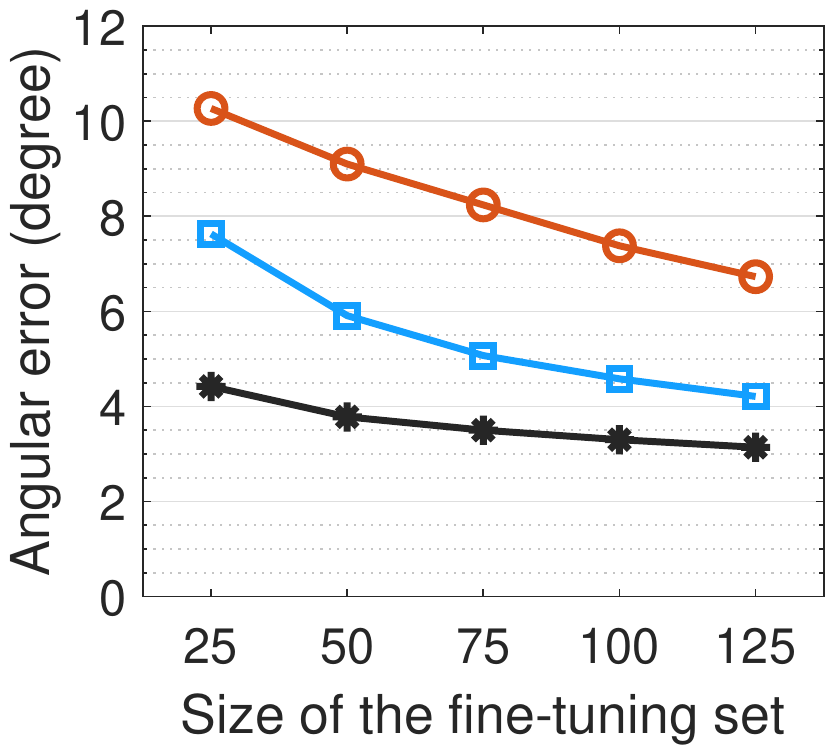}} 
	\caption{Impact of the fine-tuning set size on the gaze estimation performance: (a) ETH-XGaze and (b) MPIIFaceGaze.}
	\label{fig:finetuneSize}
	\vspace{-0.1in}
\end{figure}

We also investigate the impact of the fine-tuning set size on the gaze estimation performance. For both ETH-XGaze and MPIIFaceGaze, we use different numbers of labeled images for the calibration. The results are shown in Figure~\ref{fig:finetuneSize}. The angular errors for all examined methods decrease with the increase of the fine-tuning set size. The accuracy gain is more prominent for RGB-Random, and less significant for RGB-Supervised. With 125 samples for the fine-tuning, the performance gain between \SystemName and RGB-Supervised can decrease to 0.99$^\circ$ and 1.07$^\circ$ on ETH-XGaze and MPIIFaceGaze, respectively. 

\subsubsection{Impact of model size}
\label{subsubsc:impactOfModelSize}

To investigate the impact of model size on the gaze estimation performance, we consider two different designs of the gaze embedding network. We denote the original architecture shown in Figure~\ref{fig:cnnArch} as \textbf{\SystemName-8CNN}, indicating that it contains eight convolutional layers in the deepest path. We further design \textbf{\SystemName-12CNN}, for which the DCT coefficients matrix of the Y component is passed through \textit{four residual blocks}, which correspond to blocks conv2 and conv3 in the original ResNet-18. This makes \SystemName-12CNN four-layers deeper than \SystemName-8CNN. Table~\ref{tab:angleErrorDifferentNetworks} compares the performance of the two designs on ETH-XGaze and MPIIFaceGaze datasets. Overall, they exhibit similar performance on ETH-XGaze, while FreeGaze-12CNN achieves a modest 0.6$^\circ$ improvement over FreeGaze-8CNN on MPIIFaceGaze.



\begin{table}[]
    \caption{Averaged angular errors (mean$\pm$std) on ETH-XGaze and MPIIFaceGaze with different model sizes.}
    \label{tab:angleErrorDifferentNetworks}
    \resizebox{2.9in}{!}{
    \begin{tabular}{c|c|c|}
    \cline{2-3}
    \multicolumn{1}{l|}{}                         & \textbf{ETH-XGaze} & \textbf{MPIIFaceGaze} \\ \hline
    \multicolumn{1}{|c|}{\textbf{FreeGaze-8CNN}} & 2.95$^\circ\pm$0.51$^\circ$        & 5.07$^\circ\pm$1.03$^\circ$ \\ \hline
    \multicolumn{1}{|c|}{\textbf{FreeGaze-12CNN}} & 2.98$^\circ\pm$0.59$^\circ$        & 4.46$^\circ\pm$1.02$^\circ$           \\ \hline
    \end{tabular}
    }
    \vspace{-0.05in}
\end{table}


\subsection{Evaluation of System Latency}
\label{subsec:timeEfficiency}

Below, we profile the system latency in calibration and inference. 
We consider three different designs of the gaze embedding network: (1) \textbf{RGB-ResNet18}, which leverages the ResNet18-based network architecture. This design is adopted by the three RGB-based baseline methods introduced in Section~\ref{subsec:methodology}; (2) \textbf{\SystemName-8CNN}; and (3) \textbf{\SystemName-12CNN}. We 
consider RGB images with resolutions $224\times224$ and $448\times448$ as the original inputs. Table~\ref{tab:glops} shows the FLOPs and the number of parameters for the three system designs, given different input image resolutions. We evaluate the system latency on a desktop installed with an Intel i7-11700KF CPU and a NVIDIA GeForce RTX 3080Ti GPU, as well as a server equipped with a NVIDIA GeForce RTX 2080Ti GPU.

\begin{table}[]
\caption{The FLOPs (in the unit of $10^9$) and the number of parameters of the three network designs given different input image resolutions.}
\resizebox{3.0in}{!}{
\begin{tabular}{c|cc|c|}
\cline{2-4}
\multicolumn{1}{l|}{\multirow{2}{*}{}}          & \multicolumn{2}{c|}{\textbf{FLOPs}}                         & \multicolumn{1}{l|}{\multirow{2}{*}{\textbf{\# parameters}}} \\ \cline{2-3}
\multicolumn{1}{l|}{}                           & \multicolumn{1}{l|}{224*224} & \multicolumn{1}{l|}{448*448} & \multicolumn{1}{l|}{}                                        \\ \hline 
\multicolumn{1}{|c|}{\textbf{RGB-ResNet18}}     & \multicolumn{1}{c|}{1.82}    & 7.25                         & 11,190,912                                                   \\ \hline
\multicolumn{1}{|c|}{\textbf{FreeGaze-8CNN}}  & \multicolumn{1}{c|}{0.49}    & 1.96                         & 8,675,824                                                    \\ \hline
\multicolumn{1}{|c|}{\textbf{FreeGaze-12CNN}} & \multicolumn{1}{c|}{0.61}    & 2.42                         & 9,046,380                                                    \\ \hline
\end{tabular}
}
\vspace{-0.1in}
\label{tab:glops}
\end{table}

\subsubsection{Calibration Latency}
\label{subsubsec:calibrationTime}

We first measure the calibration latency. Specifically, we train the gaze embedding network and the gaze estimator for 1,500 epochs. The size of the fine-tuning and validation sets is 75 and 25, respectively. We repeat this procedure for five times, and report the averaged latency. The results are shown in Table~\ref{tab:calibrationLatency}. Given different settings, \SystemName-8CNN achieves 3.33 to 6.81 times speedup over the RGB-ResNet18 based design. 
Clearly, one can use a smaller fine-tuning set and a smaller training epoch to accelerate the calibration process. However, this is at the cost of sacrificing the estimation accuracy, i.e., smaller fine-tuning sets lead to higher angular error and smaller training epoch leads to network underfitting.

\begin{table}[]
\caption{Calibration latency (in min) for the three network designs with different input image resolutions. Overall, \SystemName-8CNN achieves 3.33 to 6.81 times speedup over the RGB-ResNet18 based design.}
\resizebox{3.in}{!}{
\begin{tabular}{l|cc|cc|}
\cline{2-5}
                                              & \multicolumn{2}{c|}{224$\times$224}         & \multicolumn{2}{c|}{448$\times$448}         \\ \cline{2-5} 
                                              & \multicolumn{1}{c|}{2080Ti} & 3080Ti & \multicolumn{1}{c|}{2080Ti} & 3080Ti \\ \hline
\multicolumn{1}{|l|}{\textbf{RGB-ResNet18}}   & \multicolumn{1}{c|}{6.6}    & 3.0    & \multicolumn{1}{c|}{21.1}   & 10.2    \\ \hline
\multicolumn{1}{|l|}{\textbf{FreeGaze-8CNN}}  & \multicolumn{1}{c|}{1.7}    & 0.9     & \multicolumn{1}{c|}{3.1}    & 1.8    \\ \hline
\multicolumn{1}{|l|}{\textbf{FreeGaze-12CNN}} & \multicolumn{1}{c|}{2.1}    & 1.2     & \multicolumn{1}{c|}{4.0}    & 2.3    \\ \hline
\end{tabular}
}
\vspace{-0.05in}
\label{tab:calibrationLatency}
\end{table}


\subsubsection{Inference Latency}
\label{subsubsec:inferenceTime}
To measure the inference latency, we randomly sample an image from the testing set 
for gaze estimation. The inference process involves the gaze representations extraction performed by the gaze embedding network and the gaze estimation performed by the gaze estimator. We repeat the measurement for 5,000 times, and report the averaged latency. For DCT-based methods, we also measure the time required by the frequency domain image processing module, which takes 0.1ms and 0.8ms for image with resolution 224$\times$224 and 448$\times$448, respectively. The results are shown in Table~\ref{tab:inferencelatency}. Given different settings, \SystemName-8CNN achieves 1.59 to 1.67 times speedup over the RGB-ResNet18 based design.


\begin{table}[]
    \caption{Inference latency (in ms) for the three network designs with different input image resolutions. Overall, \SystemName-8CNN achieves 1.59 to 1.67 times speedup over the RGB-ResNet18 based design.}
    \resizebox{3in}{!}{
    \begin{tabular}{l|cc|cc|}
    \cline{2-5}
     & \multicolumn{2}{c|}{224$\times$224}         & \multicolumn{2}{c|}{448$\times$448}         \\ \cline{2-5} 
     & \multicolumn{1}{c|}{2080Ti} & 3080Ti & \multicolumn{1}{c|}{2080Ti} & 3080Ti \\ \hline
    \multicolumn{1}{|l|}{\textbf{RGB-ResNet18}}   & \multicolumn{1}{c|}{27.0} & 14.9 & \multicolumn{1}{c|}{27.0} & 15.6 \\ \hline
    \multicolumn{1}{|l|}{\textbf{FreeGaze-8CNN}}  & \multicolumn{1}{c|}{17.0} & 8.9  & \multicolumn{1}{c|}{17.0} & 9.6  \\ \hline
    \multicolumn{1}{|l|}{\textbf{FreeGaze-12CNN}} & \multicolumn{1}{c|}{21.0} & 11.8 & \multicolumn{1}{c|}{22.0} & 12.0 \\ \hline
    \end{tabular}
    }
    \vspace{-0.15in}
    \label{tab:inferencelatency}
\end{table}

\vspace{0.05in}
\noindent
\textbf{Discussion.} The reductions in calibration and inference latency (reported in Tables~\ref{tab:calibrationLatency} and~\ref{tab:inferencelatency}) are not proportional to the 
FLOPs (listed in Table~\ref{tab:glops}). For instance, when the input image is with the resolution of 448$\times$448, the FLOPs of RGB-ResNet18 is 3.7 times of that of \SystemName-8CNN, but the calibration and inference latency is 5.7 and 1.6 times, respectively. This is because \textit{using FLOPs alone} cannot precisely estimates system latency on modern computation hardware, e.g., TPUs and GPUs. The system latency is often bounded by \textit{memory access costs}, and there exists different levels of optimization on modern matrix multiplication units~\cite{bello2021revisiting,jouppi2017datacenter}. For instance, a ResNet-RS model with 1.8 times more FLOPs than EfficientNet-B6 is 2.7 times faster on a TPUv3 hardware accelerator~\cite{bello2021revisiting}. Thus, 
the actual runtime system latency is affected by the computer-to-memory-ratio, the operational intensity affects, and the other optimizations on the computation hardware. 

\subsubsection{Overall System Performance Comparison}
\label{subsubsec:overallPerCom}

\begin{figure}[]
	\centering
	\subfigure[Angular error vs. calibration latency]{\includegraphics[scale=0.62]{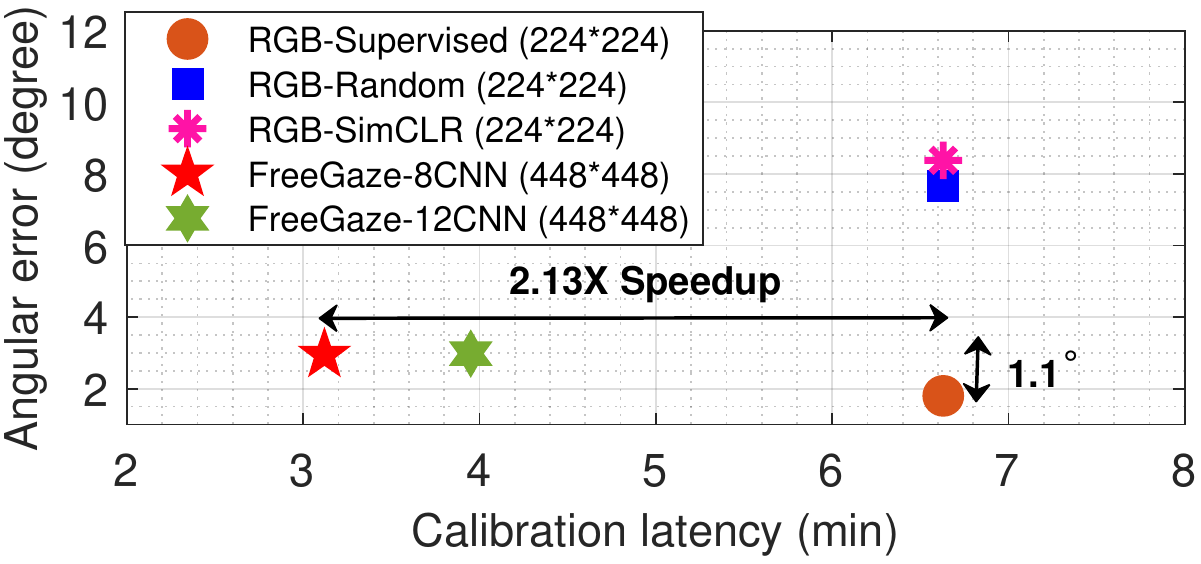}}\\[-1.5ex]
	\subfigure[Angular error vs. inference latency]{\includegraphics[scale=0.62]{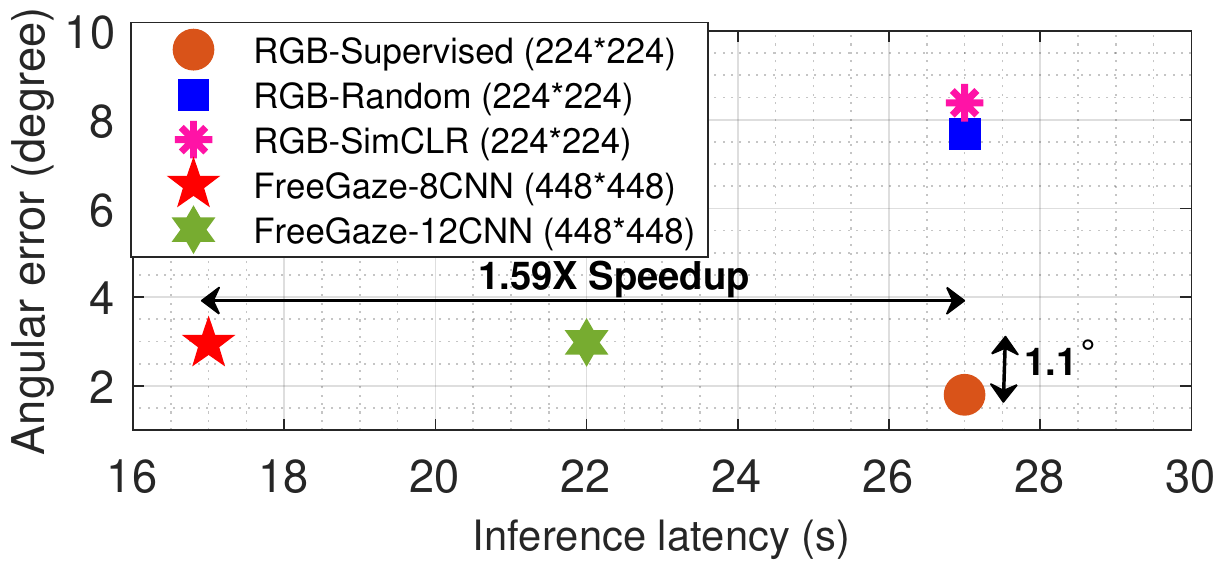}}
	\caption{Performance comparison when trading-off estimation accuracy with (a) calibration latency and (b) inference latency. 
	Comparing with RGB-Supervised, \SystemName-8CNN achieves 2.13 and 1.59 times speedup in calibration and inference, while sacrificing 1.1$^\circ$ estimation accuracy. 
	}
	\label{fig:tradeoffClibration}
	\vspace{-0.15in} 
\end{figure}

Lastly, we trade-off different design choices and perform an overall performance comparison. 
Specifically, for \SystemName-8CNN and \SystemName-12CNN, we trade-off system latency for estimation accuracy, and use images with the resolution of $448\times448$ to fine-tune the network. We do not apply this to RGB-ResNet18, as it results in 10 to 21 minutes latency for the \textit{{on the fly}} system calibration and makes it an impractical design choice. Figure~\ref{fig:tradeoffClibration} shows the overall performance comparison 
when trading-off gaze estimation accuracy (angular error) with calibration or inference latency. The results are based on the profiling on a RTX 2080Ti GPU. For the interest of space, we omit the comparison on RTX 3080Ti and the MPIIFaceGaze dataset as they exhibit similar trends. When comparing with RGB-Supervised, \SystemName-8CNN achieves 2.13 and 1.59 times speedup in calibration and inference, while sacrificing 1.1$^\circ$ estimation accuracy. 
Note that the performance gain achieved by RGB-Supervised is also at the cost of labour of intensive gaze data labeling, i.e., \textit{37K labeled images and 30K labeled images} for ETH-XGaze and MPIIFaceGaze, respectively. By contrast, \SystemName-8CNN overcomes the data labeling hurdle and achieves satisfactory gaze estimation performance in an unsupervised way.

\section{Conclusion}
\label{sec:conlusion} 

In this work, we present \SystemName, a resource-efficient framework that incorporates \textit{the frequency domain gaze estimation} and \textit{the contrastive gaze representation learning} for unsupervised gaze estimation. Our evaluation on two gaze estimation datasets demonstrates the validity of \SystemName. Specifically, \SystemName achieves comparable gaze estimation accuracy with RGB supervised learning-based approach, and reduces the angular error of existing unsupervised approach by 5.4$^\circ$ and 4.6$^\circ$ on average over the two datasets, respectively. \SystemName also enables up to 6.81 and 1.67 times speedup in system calibration and gaze estimation, respectively. 


\balance
	
\bibliographystyle{IEEEtran}
\bibliography{main}

\end{document}